\documentclass[english]{article}
\usepackage[T1]{fontenc}
\usepackage[utf8]{inputenc}
\usepackage{babel}
\usepackage{verbatim}
\usepackage{prettyref}
\usepackage{float}
\usepackage{units}
\usepackage{mathtools}
\usepackage{bm}
\usepackage{dsfont}
\usepackage{hepparticles}
\usepackage{hepnames}
\usepackage{amsbsy}
\usepackage{amstext}
\usepackage{graphicx}
\usepackage[authoryear]{natbib}
\usepackage[]
 {hyperref}

\makeatletter

\newcommand{\lyxmathsym}[1]{\ifmmode\begingroup\def\b@ld{bold}
  \text{\ifx\math@version\b@ld\bfseries\fi#1}\endgroup\else#1\fi}

\usepackage[preprint]{icml2026}

\usepackage{amsmath}
\usepackage{amssymb}
\usepackage{amsthm}

\newrefformat{cap}{\hyperref[#1]{Fig.~\ref{#1}}}
\newrefformat{fig}{\hyperref[#1]{Fig.~\ref{#1}}}
\newrefformat{tab}{\hyperref[#1]{Table ~\ref{#1}}}
\newrefformat{sec}{\hyperref[#1]{Section~\ref{#1}}}
\newrefformat{subsec}{\hyperref[#1]{Section~\ref{#1}}}
\newrefformat{sub}{\hyperref[#1]{Section~\ref{#1}}}
\newrefformat{cha}{\hyperref[#1]{Chapter~\ref{#1}}}
\newrefformat{app}{\hyperref[#1]{Appendix~\ref{#1}}}

\usepackage{bm}
\usepackage{units}

\usepackage{babel}
\usepackage{physics}
\usepackage{comment}
\usepackage{url}            %
\usepackage{booktabs}       %
\usepackage{amsfonts}       %
\usepackage{nicefrac}       %
\usepackage{microtype}      %
\usepackage{xcolor}         %
\usepackage{adjustbox}
\usepackage{stmaryrd} %

\usepackage{chngcntr}
\usepackage{makecell}

\usepackage{float}
\usepackage{bm}
\usepackage{graphicx}
\usepackage[inkscapelatex=false]{svg}
\usepackage{kantlipsum}
\usepackage{nicefrac}
\usepackage{dsfont}
\usepackage[labelfont=bf, format=plain]{caption}
\usepackage{wrapfig}

\usepackage{scalerel} %

\newlength\bbheight
\NewDocumentCommand{\myPfi}{}{%
  \text{%
    \normalfont
    \bbheight=\fontcharht\font`0 %
    \raisebox{-0.04\bbheight}[\bbheight][0pt]{%
      \includeinkscape[height=1.08\bbheight]{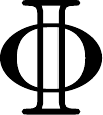}%
    }%
  }%
}

\NewDocumentCommand{\myDelta}{}{%
  \text{%
    \normalfont
    \bbheight=\fontcharht\font`0 %
    \raisebox{-0.04\bbheight}[\bbheight][0pt]{%
      \includeinkscape[height=1.08\bbheight]{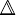}%
    }%
  }%
}

\global\long\def\dt{\dv{}{t}}%
\global\long\def\bw{\bm{w}}%
\global\long\def\NTK{\mathsf{NTK}}%

\newcommand\mylowprime{\mkern-0.5mu%
\raise0.4ex\hbox{$\scriptscriptstyle\prime$}
\mkern5mu}

\newcommand{\myshortminus}{\text{-}}
\DeclareRobustCommand{\myvarshortminus}{\scalebox{0.365}[1.0]{$-$}}
\newcommand\myinv[1]{#1^{\!\raisebox{0.3ex}{$\scriptscriptstyle\myvarshortminus$}\raisebox{0.5ex}{$\scriptscriptstyle1$}}}

\makeatletter
\DeclareRobustCommand{\T}{%
  {\mathpalette\@transpose{}}%
}
\newcommand*{\@transpose}[2]{%
  \raisebox{\depth}{$\m@th#1\intercal$}%
}
\makeatother

\newsavebox{\firstbox}
\newsavebox{\secondbox}
\newcommand{\OverlapSymbols}[3]{%
  \sbox{\firstbox}{$\displaystyle #1$}%
  \sbox{\secondbox}{$\displaystyle #2$}%
  \mathchoice
    {\ooalign{$\displaystyle #1$\cr\kern#3\wd\firstbox$\displaystyle #2$\cr}}
    {\ooalign{$#1$\cr\kern#3\wd\firstbox$#2$\cr}}
    {\ooalign{$\scriptstyle #1$\cr\kern#3\wd\firstbox$\scriptstyle #2$\cr}}
    {\ooalign{$\scriptscriptstyle #1$\cr\kern#3\wd\firstbox$\scriptscriptstyle #2$\cr}}
}

\definecolor{tabred}{RGB}{214,39,40}

\newcommand{\smallarg}[1]{\vcenter{\hbox{$\scriptstyle{#1}$}}}

\newcommand{\mydldots}{.\kern-0.1em.\kern-0.1em.}

\makeatletter
\newsavebox{\@brx}
\newcommand{\llangle}[1][]{\savebox{\@brx}{\(\m@th{#1\langle}\)}%
  \mathopen{\copy\@brx\kern-0.5\wd\@brx\usebox{\@brx}}}
\newcommand{\rrangle}[1][]{\savebox{\@brx}{\(\m@th{#1\rangle}\)}%
  \mathclose{\copy\@brx\kern-0.5\wd\@brx\usebox{\@brx}}}
\makeatother

\usepackage{accents}
\newcommand{\mydbltilde}[1]{\accentset{\approx}{#1}}

\usepackage{ulem}
\usepackage[dvipsnames]{xcolor}
\usepackage{soul}
\definecolor{Seagreen}{rgb}{0.18, 0.55, 0.34}
\definecolor{Orchid}{rgb}{0.855, 0.439, 0.839}
\definecolor{Burgundy}{rgb}{0.5, 0.0, 0.09}

\newcounter{showchanges}
\makeatother

\begin{document}
\twocolumn[
\icmltitle{A unified theory of feature learning in RNNs and DNNs}

\begin{icmlauthorlist}
\icmlauthor{Jan P. Bauer}{gatsby}
\icmlauthor{Kirsten Fischer}{juelich}
\icmlauthor{Moritz Helias}{juelich}
\icmlauthor{Agostina Palmigiano}{gatsby}
\end{icmlauthorlist}

\icmlaffiliation{gatsby}{Gatsby Computational Neuroscience Unit, UCL, London, UK}
\icmlaffiliation{juelich}{IAS-6, Forschungszentrum Juelich, Juelich, Germany}

\icmlcorrespondingauthor{Jan P. Bauer, Agostina Palmigiano}{\{j.bauer, a.palmigiano\}@ucl.ac.uk}

\icmlkeywords{RNN, DNN, Feature Learning, Mean-field Theory}

\vskip 0.3in
]

\printAffiliationsAndNotice{}
\begin{abstract}
Recurrent and deep neural networks (RNNs/DNNs) are cornerstone architectures
in machine learning. Remarkably, RNNs differ from DNNs only by weight
sharing, as can be shown through unrolling in time. How does this
structural similarity fit with the distinct functional properties
these networks exhibit? To address this question, we here develop
a unified mean-field theory for RNNs and DNNs in terms of representational
kernels, describing fully trained networks in the feature learning
($\mu$P) regime. This theory casts training as Bayesian inference
over sequences and patterns, directly revealing the functional implications
induced by the RNNs' weight sharing. In DNN-typical tasks, we identify
a phase transition when the learning signal overcomes the noise due
to randomness in the weights: below this threshold, RNNs and DNNs
behave identically; above it, only RNNs develop correlated representations
across timesteps. For sequential tasks, the RNNs' weight sharing furthermore
induces an inductive bias that aids generalization by interpolating
unsupervised time steps. Overall, our theory offers a way to connect
architectural structure to functional biases.
\end{abstract}

\section{Introduction}

\global\long\def\Pfi{\myPfi}%
\global\long\def\tPfi{\tilde{\Pfi}}%
\global\long\def\HH{\mathbb{H}}%
\global\long\def\GG{\mathbb{G}}%
\global\long\def\tHH{\tilde{\HH}}%
\global\long\def\ZZ{Z}%
\global\long\def\tZZ{\tilde{Z}}%
\global\long\def\tGG{\tilde{\GG}}%
\global\long\def\XX{\mathbb{X}}%
\global\long\def\tXX{\tilde{\XX}}%
\global\long\def\YY{\mathbb{Y}}%
\global\long\def\tYY{\tilde{\YY}}%
\global\long\def\tD{\tilde{\Delta}}%
\global\long\def\DD{\myDelta}%
\global\long\def\tDD{\tilde{\DD}}%
\global\long\def\bS{S}%
\global\long\def\bF{F}%
\global\long\def\bI{\mathbb{I}}%

\global\long\def\th{\tilde{h}}%
\global\long\def\tf{\tilde{f}}%
\global\long\def\ty{\tilde{y}}%
\global\long\def\tby{\tilde{\bm{y}}}%
\global\long\def\tz{\tilde{z}}%
\global\long\def\tbz{\tilde{\bm{z}}}%
\global\long\def\tkappa{\tilde{\kappa}}%
\global\long\def\finv#1{\frac{1}{{#1}}}%
\global\long\def\inv#1{\myinv{(#1)}}%
\global\long\def\rinv#1{\myinv{#1}}%
\global\long\def\vinv#1{\left\{  #1\right\}  ^{{\scriptstyle -1}}}%
\global\long\def\l{\mathcal{\ell}}%
\global\long\def\gry#1{{\color{gray}#1}}%
\global\long\def\red#1{{\color{red}#1}}%
\global\long\def\grn#1{{\color{green }#1}}%
\global\long\def\blue#1{{\color{blue}#1}}%
\global\long\def\bh{\boldsymbol{h}}%
\global\long\def\bbf{\boldsymbol{f}}%
\global\long\def\by{\boldsymbol{y}}%
\global\long\def\bg{\boldsymbol{g}}%
\global\long\def\bphi{\boldsymbol{\phi}}%
\global\long\def\bpsi{\bm{\psi}}%
\global\long\def\bx{\boldsymbol{x}}%
\global\long\def\bchi{\bm{\chi}}%
\global\long\def\bxi{\boldsymbol{\xi}}%
\global\long\def\bu{\boldsymbol{u}}%
\global\long\def\bv{\boldsymbol{v}}%
\global\long\def\bw{\boldsymbol{w}}%
\global\long\def\tbxi{\tilde{\boldsymbol{\xi}}}%
\global\long\def\txi{\tilde{\xi}}%
\global\long\def\tbchi{\tilde{\boldsymbol{\chi}}}%
\global\long\def\tchi{\tilde{\chi}}%
\global\long\def\tbx{\tilde{\boldsymbol{x}}}%
\global\long\def\tbw{\tilde{\boldsymbol{w}}}%
\global\long\def\tbh{\tilde{\boldsymbol{h}}}%
\global\long\def\tphi{\tilde{\phi}}%
\global\long\def\tw{\tilde{w}}%
\global\long\def\cT{\mathcal{T}}%
\global\long\def\ba{\bm{a}}%
\global\long\def\btheta{\bm{\theta}}%
\global\long\def\tba{\tilde{\bm{a}}}%
\global\long\def\ta{\tilde{a}}%
\global\long\def\tbphi{\tilde{\boldsymbol{\phi}}}%
\global\long\def\tpsi{\tilde{\psi}}%
\global\long\def\tbpsi{\tilde{\boldsymbol{\psi}}}%
\global\long\def\tx{\tilde{x}}%
\global\long\def\tbx{\tilde{\boldsymbol{x}}}%
\global\long\def\NTK{K^{\text{NTK}}}%
\global\long\def\NTKl{K^{\text{NTK},\l}}%
\global\long\def\T{\intercal}%
\global\long\def\tr{\text{tr}}%
\global\long\def\dphi{\dot{\phi}}%
\global\long\def\MMD{\text{MMD}}%
\global\long\def\RNN{\text{RNN}}%

\global\long\def\del{\partial}%
\global\long\def\delt{\partial_{t}}%
\global\long\def\dt{\frac{d}{dt}}%
\global\long\def\ds{\frac{d}{ds}}%
\global\long\def\dd#1#2{\frac{d#1}{d#2}}%
\global\long\def\ev#1{\langle#1\rangle}%
\global\long\def\vev#1{\langle#1\rangle}%
\global\long\def\cum#1{\llangle#1\rrangle}%
\global\long\def\Cum#1{\llangle#1\rrangle}%
\global\long\def\rsp#1#2{\nicefrac{#1}{#2}}%
\global\long\def\trsp#1#2{\tilde{\nicefrac{#1}{#2}}}%
\global\long\def\diag#1{\text{diag}(#1)}%
\global\long\def\mdiag#1{\bigl\llbracket#1\bigr\rrbracket}%
\global\long\def\Ev#1{\left\langle #1\right\rangle }%
\global\long\def\vEv#1{\left[#1\right]}%
\global\long\def\dbltilde#1{\mydbltilde{#1}}%
\global\long\def\det#1{|#1|}%
\global\long\def\lndet#1{\ln\,\bigl|#1\bigr|}%
\global\long\def\LLndet#1{\ln\,\Bigl|#1\Bigr|}%
\global\long\def\iid{\text{i.i.d.}}%
\global\long\def\Lndet#1{\ln\,\left|#1\right|}%
\global\long\def\dldots{\mydldots}%
\global\long\def\sN{\sqrt{N}}%
\global\long\def\is{\nu}%
\global\long\def\smlplus{{{}_+}}%
\global\long\def\smlmin{_{-}}%
\global\long\def\shortmin{\myshortminus}%
\global\long\def\tp{t\smlplus}%
\global\long\def\tm{t\smlmin}%
\global\long\def\lowprime{\mylowprime}%
\global\long\def\Tp{T^{+}}%
\global\long\def\Tm{T_{\smash{\raisebox{-0.1ex}{\myvarshortminus}}}}%
\global\long\def\sp{s\smlplus}%
\global\long\def\Tp{T\smlplus}%
\global\long\def\Sp{S\smlplus}%
\global\long\def\indep{\perp\!\!\!\!\perp}%
\global\long\def\GP{\mathcal{GP}}%
\global\long\def\N{\mathcal{N}}%
\global\long\def\L{\mathcal{L}}%
\global\long\def\O{\mathcal{O}}%

\global\long\def\ttwo{{\scriptscriptstyle 2}}%
\global\long\def\hlf{\tfrac{1}{2}}%
\global\long\def\fN{\tfrac{1}{N}}%
\global\long\def\thlf{\tfrac{1}{2}}%
\global\long\def\ttwo{{\scriptscriptstyle 2}}%
\global\long\def\shlf#1{\tfrac{#1}{2}}%
\global\long\def\vdel{\not\partial_{t}}%
\global\long\def\D{\mathcal{D}}%
\global\long\def\N{\mathcal{N}}%
\global\long\def\order{\mathcal{O}}%
\global\long\def\bR{\mathbb{R}}%
\global\long\def\I{\mathbb{I}}%
\global\long\def\kap{{\scriptstyle \kappa}}%
\global\long\def\Kap{\mathsf{K}}%
\global\long\def\sdelta{{\scriptstyle \delta}}%
\global\long\def\bsdelta{{\scriptstyle \bar{\delta}}}%

\global\long\def\order{\mathcal{O}}%
\global\long\def\cW{\mathcal{W}}%
\global\long\def\th{\tilde{h}}%
\global\long\def\I{\mathbb{I}}%
\global\long\def\cR{\mathbb{R}}%
\global\long\def\ty{\tilde{y}}%
\global\long\def\cZ{\mathcal{Z}}%
\global\long\def\tr{\mathrm{tr}}%
\global\long\def\const{\mathrm{const}}%
\global\long\def\erf{\mathrm{erf}}%
\global\long\def\tf{\tilde{f}}%
\global\long\def\i{{\scriptstyle \iota}}%
\global\long\def\cov{\text{Cov}}%
\global\long\def\var{\text{Var}}%
\global\long\def\std{\text{std}}%

\global\long\def\gw{{\scriptstyle w}}%
\global\long\def\gu{{\scriptstyle u}}%
\global\long\def\gv{{\scriptstyle v}}%
\global\long\def\v{{\scriptstyle v}}%
\global\long\def\gxi{{\scriptstyle \sigma}}%
\global\long\def\ggxi{{\scriptstyle \gxi^{\ttwo}}}%
\global\long\def\gx{{\scriptstyle xx}}%
\global\long\def\ggx{{\scriptstyle \gx}}%
\global\long\def\w{{\scriptstyle \gw}}%
\global\long\def\u{\gu}%
\global\long\def\gy{{\scriptstyle y}}%
\global\long\def\bW{\bm{W}}%
\global\long\def\bU{\bm{U}}%
\global\long\def\gy{{\scriptstyle y}}%
\global\long\def\ggy{\gy^{\ttwo}}%
\global\long\def\bJ{\bm{J}}%
\global\long\def\gj{{\scriptstyle j}}%
\global\long\def\ggj{\gj^{\ttwo}}%
\global\long\def\bV{\bm{V}}%
\global\long\def\tvctr#1{\smallarg{#1}}%
\global\long\def\ovrst#1#2{\overset{#2}{#1}}%

\begin{figure}[t]
\centering
\centering{}\includegraphics[width=0.9\columnwidth]{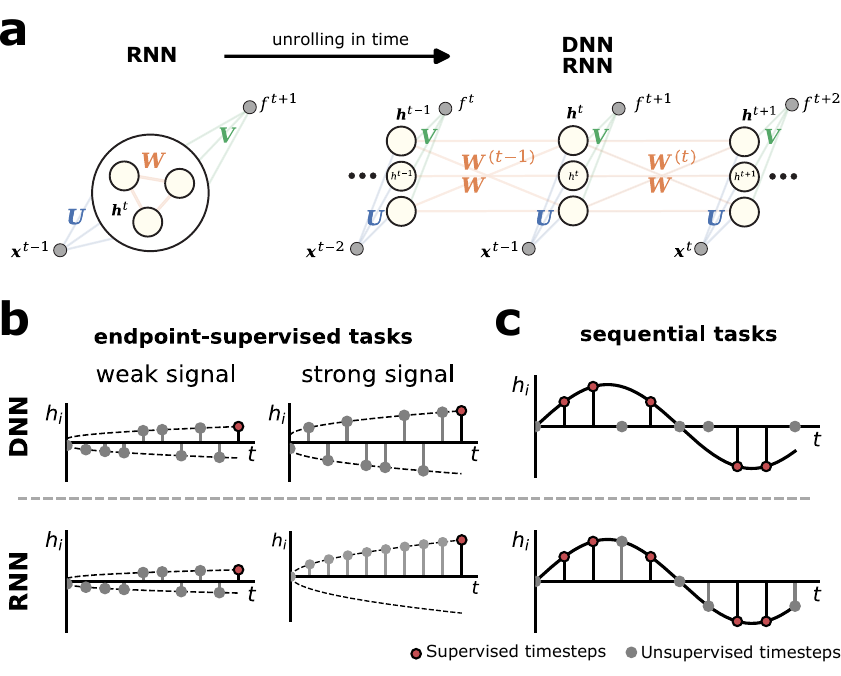}\caption{\textbf{Graphical abstract: RNNs resemble DNNs after unrolling-in-time,
but differ functionally depending on signal strength and tasks. a)}
Recurrent neural network (RNN, left) and its unrolling-in-time representation
(right): the same recurrent weights $\protect\bW$ are shared across
timesteps. \textbf{b)}\textit{ Endpoint-supervised tasks}: Supervision
target at the last layer (red) affects hidden layer representation
(gray), but only in the RNN and for sufficient signal strength induces
a phase transition towards temporal coherence. \textbf{c)} \textit{Sequential
tasks}: The RNNs weight sharing induces a temporally coherent inductive
bias that facilitates generalization from supervised (red) to unsupervised
timepoints (gray), whereas the DNN exhibits \textquotedblleft regression
to the mean\textquotedblright{} for unsupervised points due to an effectively
white prior.}\label{fig:graphical_abstract}
\end{figure}
Understanding how neural networks learn useful representations from
data remains a central question in machine learning theory. While
recent work has studied feature learning in DNNs \citep{Yang20_14522,bordelon2022a,Cui23_6468},
the corresponding theory for RNNs is far less developed. RNNs process
sequences by repeatedly applying the same transformation over time,
so that each new hidden state is computed using the same set of recurrent
weights $\bW$. If we “unroll” this computation in time, the RNN can
be viewed as a deep feedforward network whose layers correspond to
timesteps. From this view, the two network architectures appear remarkably
similar (\prettyref{fig:graphical_abstract}\textbf{)}. However, a
difference is that the weights are shared across all layers (timesteps)
in the RNN, whereas a standard DNN has layer-specific weights. It
is so far unknown how weight sharing affects computation: are the
extra, untied parameters of a DNN always advantageous, or can the
inductive bias imposed by weight sharing actually improve learning
by better aligning the model with the structure of a task?

Our key contributions towards this question are:
\begin{itemize}
\item We develop a unified mean-field theory for RNNs and DNNs in terms
of representational kernels, describing trained networks after convergence
of Langevin stochastic gradient descent in the feature learning ($\mu$P)
regime. This theory casts network training as Bayesian inference over
timesteps and patterns, revealing the functional implications of their
architectural structure.
\item For tasks that are typical for DNNs (``\textit{endpoint-supervised}''
tasks), we analytically show that even within $\mu$P scaling, different
phases of feature learning exist. Below a critical signal strength,
the representational kernels of RNNs and DNNs coincide, although they
differ from the kernels obtained in a lazy learning phase (NNGP),
which characterize initialization. Beyond a critical signal strength,
however, the RNN's representation enters a \textit{temporally coherent}
phase that is absent in the DNN.
\item For tasks that are typical for RNNs (``\textit{sequential}'' tasks),
we show how the RNN's weight sharing induces an inductive bias that
enables more sample-efficient learning. This implies a functional
advantage beyond the representational changes of the endpoint-supervised
case, and underlines that task-model alignment not only depends on
expressivity, but also on adaptivity to task structure.
\end{itemize}
The code to reproduce all figures is available at \href{http://github.com/japhba/temporal-feature-learning}{github.com/japhba/temporal-feature-learning}.

\section{Related work}

\paragraph{Probabilistic descriptions of neural networks.}

The study of feedforward neural networks in the infinite-width limit
has a rich history, beginning with the connection to Bayesian inference~\citep{Neal96,Lee18}.
These foundational works established that infinitely wide neural networks
with random weights under the standard parametrization (SP, all weights
scale as $1/\sqrt{\text{width}}$) converge to Gaussian processes.
More recently, this line of work has been generalized using mean-field
methods to account for feature learning either by adopting $\mu$P
scaling (\citet{Yang20_14522}, readout weights scale as $1/\text{width}$),
or by considering proportional limits of width and number of data
samples. Such theories have been formulated in terms of adaptive weight
scales \citeauthor{Li21_031059,Ariosto2022,pacelli2023}, task-aligned
readout weights \citep{vanmeegen2025}, or kernel matrices \citep{Seroussi21Separationscalesthermodynamic,ZavatoneVeth21_NeurIPS_I,Fischer24_10761,lauditi2025},
providing a probabilistic description of representations that goes
beyond dynamics near initialization as described by the neural tangent
kernel (NTK) \citep{Jacot18_8580}. We here develop a generalization
of such kernel theories to RNNs in a way that treats RNNs and DNNs
on equal footing.

\paragraph{Bayesian inference in state-space models.}

Our approach to analyzing recurrent networks directly connects to
Bayesian inference in classical probabilistic models. In particular,
for linear activation, the RNN architecture we consider is identical
to a linear Gaussian state-space model with vanishing noise in the
forward pass \citep{kalman1960new}. Whereas these models consider
inference at fixed weights via Kalman filtering, we here consider
a posterior over weights due to training as well, leading to a more
general theory.

\paragraph{Theory of recurrent neural networks.}

The dynamics of randomly-coupled recurrent neural networks have been
studied extensively. Dynamical mean field theory has been derived
in networks with random unstructured \citep{Sompolinsky88Chaosrandomneural,Molgedey92Suppressingchaosneural,Toyoizumi11edgechaosAmplification},
excitatory and inhibitory \citep{Kadmon15Transitionchaosrandom,Sanzeni_Palmigiano_2023,Mastrogiuseppe17_e1005498}
and low-rank \citep{Mastrogiuseppe18_1139,Landau18_e1006309} connectivity
\citep{Schuecker18_041029,Segadlo22_accepted,schuessler2024}, including
their response to perturbations \citep{Sanzeni_Palmigiano_2023,Nguyen_2025,Palmigiano_2023}
and computational properties (such as memory \citep{Toyoizumi11_051908,Schuecker18_041029,pereira2023forgetting}).
The study of learning in RNNs is less thoroughly explored. The recurrent
neural tangent kernel (RNTK) \citep{alemohammad2021} extends the
NTK framework \citep{Jacot18_8580} to recurrent architectures, providing
insights into their training dynamics in the infinite-width lazy training
regime. \citet{proca2025,bordelon2025} analyzed the learning \textit{dynamics}
in recurrent linear networks in the balanced regime. \citet{bordelon2024b}
presented ideas to study learning as Bayesian inference in RNNs in
continuous time with a leak term \citep{Amari72_643}.

\paragraph{Generalization, architectural inductive biases and feature learning.}

Kernel methods have been employed to study generalization \citep{canatar2021,Simon2023,vanmeegen2025,rubin2025},
with tight links to the training of neural networks and the role of
inductive biases \citep{aiudi2025}. Meanwhile, other work has explored
the impact of initialization necessary for feature learning, in particular
the role of small initialization \citep{Saxe13Exactsolutionsnonlinear,atanasov2021,kunin2024,tu2024}.
We here study feature learning in terms of structured updates to the
kernel's eigenvectors, highlighting the effects of architecture and
initialization.

\paragraph{Learning dynamics versus convergence.}

Previous work has studied the \textit{dynamics} of learning \citep{Saxe14_iclr,bordelon2025,proca2025},
by approaches including \textit{dynamical} mean-field theory \citep{bordelon2022a},
but comes at the cost of making simplifying assumptions \citep{Saxe14_iclr,proca2025}
or complicated expressions. We here join a line of recent work that
instead seeks out to only describe the network state after training
when the weights have converged to a stationary distribution, significantly
simplifying the theory \citep{Seung1992,Cohen21_023034,seroussi23_908,Cui23_6468,Fischer24_10761,vanmeegen2025,lauditi2025}.

\section{Results}

\global\long\def\tPfi{\tilde{C}}%

\subsection{Unified feature learning theory for RNNs and DNNs}\label{subsec:MFT-main}

We consider a general architecture with hidden dimension $N$ and
input dimension $D$. At each timestep $t$, the pre-activations $\bh^{t}\in\mathbb{R}^{N}$
receive external temporally-dependent input $\bx^{t-1}\in\mathbb{R}^{D}$
through read-in weights $\bU\in\mathbb{R}^{N\times D}$ and recurrent
inputs $\phi(\bh^{t-1})$ through hidden weights $\bW\in\mathbb{R}^{N\times N}$,
where $\phi(\circ)$ is an elementwise activation function. The scalar
output of the architecture $f^{t+1}$ is obtained through readout
weights $\bV\in\mathbb{R}^{1\times N}$, so that the whole architecture
reads
\begin{eqnarray}
\bh^{t} & = & \bW^{(t-1)}\phi(\bh^{t-1})+\bU\bx^{t-1}\,,\label{eq:weight-model-RNN}\\
f^{t+1} & = & \bV\phi(\bh^{t})\,,\nonumber \\
\nonumber \\\bW^{(t)} & = & \begin{cases}
\bW, & \text{RNN}\\
\bW^{(t)}, & \text{DNN}
\end{cases}\nonumber 
\end{eqnarray}
with $t=1,\ldots,T-1$. We will abbreviate this penultimate timestep
with $\Tm\coloneqq T-1$ from here on.\\
We group all parameters into $\Theta$ and abbreviate $\bphi^{t}\coloneqq\phi(\bh^{t})$.
Besides RNNs, this definition encompasses also DNNs, where $\bx^{0}$
is the input and $f^{T}$ the output (see \prettyref{subsec:MFT}
for details).\\
We optimize this model via gradient descent on a mean-squared loss
$\L(\Theta;y,\bx)=\hlf\frac{1}{P|\mathcal{T}|}\sum_{p}^{P}\sum_{t\in\mathcal{T}}\,(y_{p}^{t}-f_{p}^{t}(\Theta,\bx))^{2}$
over samples $p$ and supervised timesteps $t\in\mathcal{T}$. For
each gradient update, we allow for an i.i.d. Gaussian noise of strength
$\sqrt{2\Kap}$ and an independent weight decay of strength ${\scriptstyle \frac{\Kap}{G_{\bm{\theta}}}}$,
so that the overall update reads 
\begin{equation}
\bm{\theta}_{s+1}=\bm{\theta}_{s}-\nabla_{\bm{\theta}}{\scriptstyle PT}\L(\Theta_{s})\,\mathrm{d}s-{\scriptstyle \frac{\Kap}{G_{\bm{\theta}}}}\bm{\theta}_{s}\,\mathrm{d}s+{\scriptstyle \sqrt{2\Kap\,\mathrm{d}s}\,}\bxi_{s}\,.\label{eq:SGD}
\end{equation}
This update is commonly referred to as SGLD (stochastic gradient Langevin
dynamics) \citep{Naveh20_01190}. This can be thought of as a naive
approximation of stochastic gradient descent (SGD), replacing mini-batch
fluctuations by uncorrelated noise. It can be shown that after convergence
this algorithm samples from a stationary distribution $\bm{\theta}\sim P(\Theta|y,\bx)\propto\exp\{-{\scriptstyle PT}\L(\Theta;y,\bx){\scriptstyle /\Kap}\,-\,\hlf\sum_{\bm{\theta}}\lVert\bm{\theta}\rVert^{2}{\scriptstyle /G_{\btheta}}\}$
\citep{Gardiner83,Seung1992,Kardar2007_Paticles}. This distribution
can likewise be interpreted as a Bayesian posterior combining two
factors: first, a base distribution that is a Gaussian i.i.d. weight
prior $P(\Theta)\text{\ensuremath{\propto}}\exp\{-\hlf\sum_{\bm{\theta}}\lVert\bm{\theta}\rVert^{2}{\scriptstyle /G_{\btheta}}\}\propto\N(\Theta|0,\,G_{\bm{\theta}})$,
and a likelihood $\propto\exp\{-{\scriptstyle PT}\L(\Theta;y,\bx){\scriptstyle /\Kap}\}$,
which may alternatively be regarded as a Bayesian regularization due
to noisy labels, $y_{p}^{t}=f_{p}^{t}+{\scriptstyle \sqrt{\Kap}}\xi_{p}^{t}$.\\
We show in \prettyref{subsec:MFT} that it is possible to translate
``structure to function'' by marginalizing out the prior weights
$\Theta$ in favor of a distribution that only depends on neural activations
$h$, input $\bx$, and labels $y$:
\begin{multline}
P(h|y,\bx)\propto\exp\Bigl\{-\hlf\tr\,[\YY_{\mathcal{T}}\inv{\v\Pfi_{\mathcal{T}}^{-}+\kap}]\\
\quad-\hlf h^{\T}\inv{\w\mdiag{\Pfi^{-}}+\u\XX^{-}}h+\phi(h)^{\T}\,\tPfi\,\phi(h)\Bigr\}.\label{eq:P_y__x}
\end{multline}
In making this change, this theory summarizes the effect of weight
learning in the form of a\textit{ kernel}
\begin{align}
\Pfi_{pp}^{tt'} & \coloneqq\fN\sum_{i}^{N}\phantom{\langle}\phi(h_{i,p}^{t})\phi(h_{i,p'}^{t'})\phantom{\rangle}_{\phantom{P(h|y,\bx)}}\label{eq:C}\\
 & \overset{\mathclap{N\shortrightarrow\infty}}{\asymp}\hphantom{\fN\sum_{i}^{N}}\langle\phi(h_{p}^{t})\phi(h_{p'}^{t'})\rangle_{P(h|y,\bx)}\,.\label{eq:C_MFT}
\end{align}
In these expressions, contractions $a^{\T}b\coloneqq\sum_{p}^{P}\sum_{t}^{T}a_{p}^{t}b_{p}^{t}$
are taken over the joint pattern-time space, and $(\circ^{-})^{tt^{\prime}}\coloneqq\circ^{t-1,t^{\prime}-1}$
is a shorthand to indicate time-shift. Likewise, $\XX_{pp'}^{tt'}\coloneqq\frac{1}{D}\sum_{i}^{D}x_{p,i}^{t}x_{p',i}^{t'}$
and $\YY_{pp'}^{tt'}\coloneqq y_{p}^{t}y_{p'}^{t'}$ denote input
and label kernels, and $\HH_{pp'}^{tt'}\coloneqq\fN\sum_{i}^{N}h_{p,i}^{t}h_{p',i}^{t'}\asymp\vev{h_{p}^{t}h_{p'}^{t'}}_{P(h|y,\bx)}$
is the kernel for the preactivations, defined in analogy to the kernel
$\Pfi$ for the postactivations in \prettyref{eq:C}. A subscript
$\circ_{\cT}$ denotes the restriction of a kernel to the supervised
timesteps $\cT$, and $\Pfi$, $\HH$, and $\tPfi$ are matrices in
$\mathbb{R}^{PT_{-}\times PT_{-}}$. In this theory, RNNs and DNNs
only differ by a \textit{masking operation }in time, $\mdiag{\Pfi^{-}}\coloneqq\bigl\{\begin{smallmatrix}\Pfi^{-}, & \text{RNN}\\
\text{diag}(\Pfi^{-}), & \text{DNN}
\end{smallmatrix}$, capturing the effect of the DNN's weight independence in form of
the correlation of the $h$-fields across timesteps. We discuss the
effect of this difference in the following sections.\\
As we show in \prettyref{subsec:MFT} in the large $N$ limit, the
empirical averages $\fN\sum_{i}^{N}$ concentrate to a well-defined
value. The conjugate variable $\tPfi(\HH,\Pfi)=\tPfi_{y}+\tPfi_{h}$
encodes the two constraints that the activity $h$ is subjected to:
$\tPfi_{y}=\hlf\v\inv{\v\Pfi^{-}+\kap}\YY\inv{\v\Pfi^{-}+\kap}$ aligns
the representation towards the targets $y^{t}$, whereas $\tPfi_{h}=\hlf\w\inv{\w\mdiag{\Pfi^{-}}+\u\XX^{-}}\Bigl(\HH\:-\:(\w\mdiag{\Pfi^{-}}+\u\XX^{-})\Bigr)\inv{\w\mdiag{\Pfi^{-}}+\u\XX^{-}}$
acts as a force on $\phi(h)$ towards the prior describing the forward
pass $\bh^{t}=\bW^{(t-1)}\bphi^{t-1}+\bU\bx^{t-1}$ with untrained
weights. \\
The scalars $u,w,v$ are intensive parameters that reparametrize the
scales of the weights $G_{\bm{\theta}}$ so that the kernels become
intensive (i.e., $\O(1)$ with respect to $N$) quantities that concentrate,
and we likewise reparametrized $\Kap$ by the intensive quantity $\kappa$
(see \prettyref{subsec:MFT}). In the limit of large networks, the
replacement of the empirical average over neuron indices \prettyref{eq:C}
with an average that factorizes over neurons \prettyref{eq:C_MFT}
becomes exact; that is, the empirical average is self-averaging due
to concentration.\\
In deriving \prettyref{eq:P_y__x}, we traded the posterior distribution
$P(\Theta|y,\bx)$ over parameters of extensive size in $N$ for a
distribution $P(h|y,\bx)=P(h|\Pfi,\tPfi,\YY,\XX)$ over a single variable
$h_{i}\equiv h$ which is identically distributed over neurons, and
only depends on the kernels $\Pfi,\tPfi,\YY,\XX$. This theory straightforwardly
recovers the $\mu$P scaling \citep{Yang20_14522} for the prior variances
as $(G_{\bm{\bU}},G_{\bm{\bW}},G_{\bm{\bV}})=(\nicefrac{u}{D},\nicefrac{w}{N},\nicefrac{v}{N^{2}})$
as well as $\Kap=\nicefrac{\kap}{N}$ 
, from two requirements: that pre-activations $h$ scale suitably
in the $N\to\infty$ limit (in particular avoiding exploding/vanishing
gradients), and permitting feature learning in terms of a change to
the representation kernel $\Pfi$. Moreover, the theory will allow
us to directly characterize representational and functional properties
of the networks, which will be our focus in the following.
\begin{figure}
\centering
\centering{}\includegraphics[width=1\columnwidth]{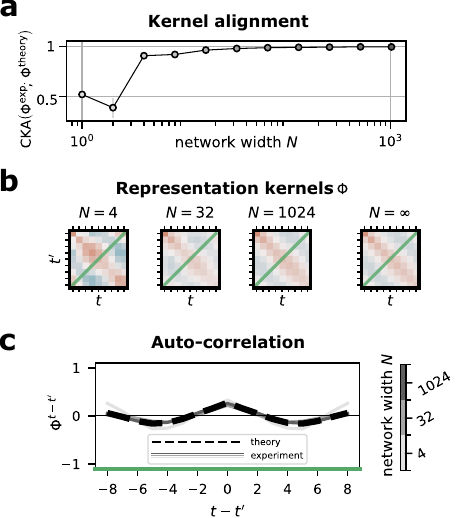}\caption{\textbf{Kernels in trained RNNs converge to theory predictions for
large network width $N$.} We train an RNN as described in \prettyref{eq:SGD}
with nonlinear activation function $\phi(\circ)=\text{erf}(\tfrac{\sqrt{\pi}}{2}\circ)$
to produce a sinusoidal target sequence $y^{t}=\cos(\frac{2\pi}{T}t)$,
$T=10$ in response to a scalar input at time $t=0$, i.e. $x^{t}=\delta^{t0}$.
\textbf{a)} Centered kernel alignment (CKA) between the kernel $\protect\Pfi_{\text{exp.}}(\Theta)=\protect\fN\sum_{i}^{N}\phi(h_{i})\phi(h_{i})^{\protect\T}$
from explicit weight SGLD experiments at different network widths
$N$ and the kernel $\protect\Pfi_{\text{theory}}=\protect\vev{\phi(h)\phi(h)^{\protect\T}}_{P(h|y,\protect\bx)}$
predicted by the theory \prettyref{eq:C_MFT} for $N\rightarrow\infty$.
\textbf{b)} Temporal structure of kernels $\protect\Pfi_{\text{exp.}}$
for different finite network widths compared to $\protect\Pfi_{\text{theory}}$
at infinite width $N\rightarrow\infty$. \textbf{c) }Autocorrelation
function of the network measured along the anti-diagonal $\protect\Pfi^{t-t'}$
of the kernel (dashed: theory; full curve: numerics), marked in green
in panel\textbf{ b.}}\label{fig:weights_match_kernels}
\end{figure}
We first test whether the theory can accurately capture the effects
of weight sharing via a simple time-series regression task in \prettyref{fig:weights_match_kernels}
before we leverage it in the following sections to explain consequences
of weight sharing. To this end, we compare kernels $\Pfi_{\text{exp.}}^{tt'}(\Theta)=\fN\sum_{i}^{N}\phi(h_{i}^{t})\phi(h_{i}^{t'})$
measured from networks trained through SGLD experiments in weight
space, and kernels $\Pfi_{\text{theory}}^{tt'}=\vev{\phi(h^{t})\phi(h^{t'})}_{P(h|y,\bx)}$
predicted by the theory \prettyref{eq:P_y__x}. We use centered kernel
alignment $\text{CKA}(\Pfi_{\text{exp.}}(\Theta),\,\Pfi_{\text{theory}})$
\citep{Cortes12_795,Fischer24_10761}, which computes the cosine similarity
of the vectorized matrices after subtracting their means. In \prettyref{fig:weights_match_kernels}\textbf{a},
the alignment between theory and simulations increases with the network
width $N$ as expected, since the empirical average we defined in
\prettyref{eq:C} converges with the number of constituting terms.
Notably, we obtain an accurate description of the network kernels
already at large but finite network width. \\
The trained networks exhibit a non-trivial temporal correlation structure,
visible as non-diagonal kernels in Fig. \prettyref{fig:weights_match_kernels}\textbf{b};
the anti-diagonal yields the autocorrelation function shown in \prettyref{fig:weights_match_kernels}\textbf{c},
which likewise converges to the theoretical prediction for large $N$.
In this setting, where there are no correlations in the input, there
are potentially two contributions to the observed temporal coherence:
the weight sharing over time and the temporal correlation in the supervision
signal $y$. In the following sections, we use our theory to separate
these contributions, showing that such temporal correlations are specific
to RNNs due to weight sharing across timesteps, which is absent in
feed-forward architectures. In \prettyref{subsec:RNN-DNN-comp}, we
focus on \textit{endpoint-supervised} tasks that are the typical use
case of DNNs, and in \prettyref{subsec:efficient-seq-learners} we
focus on \textit{sequential }tasks\textit{ }that are the typical use
case of RNNs.

\subsection{Phase transition towards temporal coherence in RNNs induced by strong
learning signal in endpoint-supervised tasks}\label{subsec:RNN-DNN-comp}

\begin{figure*}
\centering
\centering{}\includegraphics[width=1\textwidth]{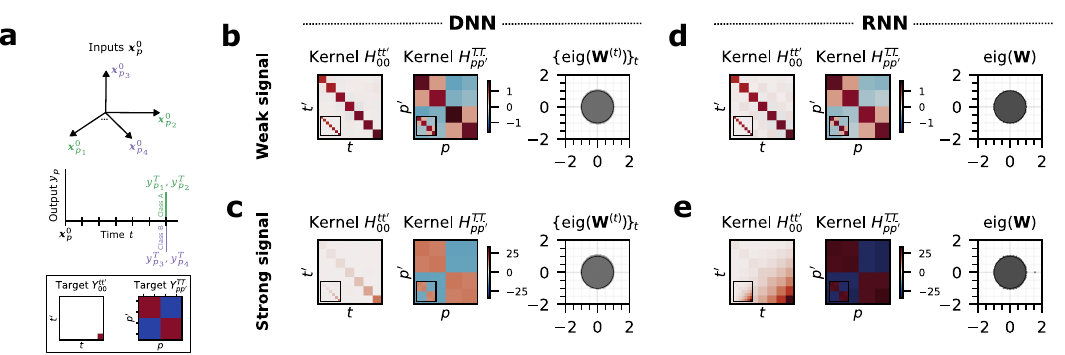}\caption{\textbf{RNNs and DNNs learn similar spatial representations while
only RNNs learn temporal coherence for strong learning signal. a)}
Binary classification task: $P=4$ pairwise orthogonal inputs $\protect\bx_{p}\in\protect\bR^{D}$
with $D=4$ map to labels $y_{p}\in\{-1,1\}$, as summarized by kernels
$\protect\YY_{00}^{tt'}\in\protect\bR^{T_{-}\times T_{-}}$, $\protect\YY_{pp'}^{TT}\in\protect\bR^{P\times P}$.
\textbf{b-e)} Kernel and weight structure of DNNs (panels \textbf{b,
c)} and RNNs (panels \textbf{d}, \textbf{e}) trained on the task.
Lower-left insets show prediction by the kernel theory. We consider
the cases of weak (upper row) or strong learning signal (lower row).
From left to right in each cell (panels \textbf{b}-\textbf{e}): temporal
kernel $\protect\HH_{00}^{tt'}$ for fixed pattern $p=0$, sample
kernel $\protect\HH_{\protect\smash{pp\protect\lowprime}}^{\protect\Tm\protect\Tm}\protect\coloneqq\protect\fN\protect\bh_{\protect\smash{p}}^{\protect\Tm}\cdot\protect\bh_{\protect\smash{p\protect\lowprime}}^{\protect\Tm}$
in last timestep $\protect\Tm\protect\coloneqq T-1=7$, and eigenspectrum
in the complex plane of hidden weights $\protect\bW$ (RNN) or $\{\protect\bW^{(t)}\}_{t}$
(DNN, eigenspectra of all $\{\protect\bW^{(t)}\}_{t}$ plotted on
same axis), with $N=2048$. Other parameters: $w=v=u=1$, $\kappa=0.1$.}\label{fig:RNNs_vs_DNNs_endpoint_sv}
\end{figure*}
The formulation in \prettyref{eq:weight-model-RNN} allows us to draw
a direct comparison to deep neural networks (DNNs) via unrolling RNNs
in time with a one-to-one correspondence between RNN timesteps $t$
and DNN layers, which we thus also index by $t$. DNNs are typically
trained on tasks in which inputs are provided only at the first layer
and supervision is applied only at the final layer; we hereafter refer
to this setting as \textit{endpoint-supervised}. Equation \prettyref{eq:weight-model-RNN}
for the DNN then becomes
\begin{align}
\bh^{1} & =\bU\bx^{0}\,,\label{eq:weight-model-DNN}\\
\bh^{t} & =\bW^{(t-1)}\phi(\bh^{t-1})\,,\quad2\le t<T\,,\nonumber \\
f^{T} & =\bV\phi(\bh^{T-1})\,.\nonumber 
\end{align}
As we detail in \prettyref{subsec:Reduction-to-DNN}, DNNs can be
understood as a special case of the general mean-field theory developed
in \prettyref{subsec:MFT-main} by omitting input at timesteps $t>1$,
and having no supervision target at timesteps $t<T$. Notably, DNNs
in addition have independent weight priors $\bW^{(t)}$ across timesteps,
reflected in a factorization of the prior introduced in \prettyref{subsec:MFT-main},
$P_{\text{DNN}}(\{\bW^{(t)}\}_{t})=\prod_{t=1}^{T-1}P(\bW^{(t)})$.

Despite these similarities, it is unclear what is the effect that
weight sharing in the RNN has on representation (i.e., the kernels
$\Pfi$, $\HH$) or on computation (i.e., the model output $f$).
\citet{Segadlo22_accepted} show that in the NNGP limit (i.e., when
non-readout weights don't change during training), temporal correlations
in RNNs with point-symmetric activation functions vanish, and the
kernel representations in DNNs and RNNs therefore coincide. Here,
we generalize their theoretical approach to account for feature learning
under $\mu$P scaling. Our theory reveals a distinction between the
kernels in RNNs and DNNs in endpoint supervised tasks, but only if
the learning signal is sufficiently strong (\prettyref{fig:RNNs_vs_DNNs_endpoint_sv},
here for simplicity for linear activation, i.e. $\phi(h)\rightarrow h$,
$\Pfi\rightarrow\HH$). We find a qualitative change in the RNN's
kernel representation from a \textit{temporally incoherent} to a \textit{temporally
coherent} regime that is absent in DNNs. Here, temporally coherent
means that the vectors $\fN\bh^{t}\cdot\bh^{t^{\prime}}$ between
timesteps $t$ and $t^{\prime}$ align. This change is accompanied
by an outlier eigenvalue in the spectrum of $\bW$. We find that this
transition is controlled by the strength of the learning signal $\lambda=\nicefrac{y^{2}}{vw^{T-2}u}$
relative to the $\O(1)$ parameters $u,w,v$ that define the scale
of the weights, a dependence that we characterize more quantitatively
below. \\
The pattern-by-pattern kernel $\HH_{pp'}^{\Tm\Tm}$ in the last time
point $\Tm\coloneqq T-1$ determines the predictor statistics of the
network via $f^{T}=\v\HH^{\Tm\Tm}\inv{\v\HH^{\Tm\Tm}+\kap}y^{T}$
\citep{Neal96,Jacot18_8580}. Interestingly, \prettyref{fig:RNNs_vs_DNNs_endpoint_sv}
reveals that the block structure induced by class membership is present
in both RNNs and DNNs and differs only in scale, hence leading to
qualitatively similar predictors. This reflects that both networks
have learned the pattern-by-pattern structure $\YY$ that defines
the task (\prettyref{fig:RNNs_vs_DNNs_endpoint_sv}\textbf{a}), in
particular the last timestep's representation $\HH_{pp'}^{\Tm\Tm}\simeq\YY=yy^{\T}$
has learned $y$ as an eigenvector (cf. \citet{Fischer24_10761}).

\begin{figure*}
\centering
\centering{}\includegraphics[width=0.8\textwidth]{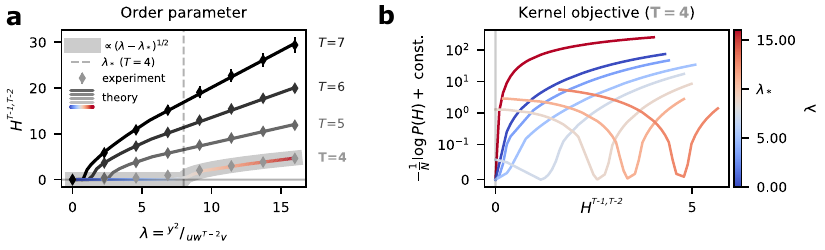}\caption{\textbf{Second-order phase transition in linear RNNs of $T=4$ with
critical exponent $\frac{1}{2}$. a)} Off-diagonal kernel order parameter
$\protect\HH^{{\scriptscriptstyle T\protect\shortmin1,T\protect\shortmin2}}$
as a function of the control variable $\lambda$. Solid curves: kernel
theory for different network depths. Diamonds: empirical kernel from
weight SGLD. \textbf{b)} Negative log-probability for the off-diagonal
order parameter $\protect\HH^{-}$, obtained from maximizing $P(\protect\HH|\protect\HH^{{\scriptscriptstyle {\scriptscriptstyle T\protect\shortmin1,T\protect\shortmin2}}},y,\protect\bx)$
as a function of the signal strength control variable $\lambda$.
Network width $N=2048$ (error bars: residual fluctuation at the equilibrium
of the update \prettyref{eq:SGD}).}\label{fig:Landau_L4}
\end{figure*}
We now seek a theoretical understanding behind this phenomenon. For
simplicity, we here consider only the linear case (we show in \prettyref{sec:Additional-figures}
that the transition also persists in the non-linear case). From the
theory \prettyref{eq:P_y__x} that yields the probability distribution
of the preactivations, we obtain a probability over representations
via $P(\HH|y,\bx)=\vev{\delta(\HH-hh^{\T})}_{P(h|\Pfi,\tPfi,\YY,\XX)}$
after replacing $\tPfi=\tPfi(\HH)$ by its stationarity condition
(\prettyref{sec:linear}):
\begin{multline}
-\ln\,P(\HH|y,\bx)/N=\hlf\tr\,[\YY_{\cT}\inv{\v\HH_{\cT}^{-}+\kap}]\\
\quad+\hlf\tr\,[\HH\inv{\w\mdiag{\HH^{-}}+\u\XX^{-}}]-\hlf\ln\frac{\det{\HH}}{\det{\w\mdiag{\HH^{-}}+\u\XX^{-}}}\,,\label{eq:P_H_lin}
\end{multline}
where $\mdiag{\HH^{-}}\coloneqq\bigl\{\begin{smallmatrix}\HH^{-}, & \text{RNN}\\
\text{diag}(\HH^{-}), & \text{DNN}
\end{smallmatrix}$ again is the architecture-dependent masking and $\det{\circ}$ is
the determinant. The second line takes the form of a Kullback-Leibler
divergence $D_{\mathrm{KL}}\!\big(\mathcal{N}_{h}(\HH)\,\|\,\mathcal{N}_{h}(\w\mdiag{\HH^{-}}+\u\XX^{-})\big)$
that vanishes at the NNGP prior which describes the forward propagation
under random weights. For the endpoint-supervised tasks we consider
here, the first term has support only on the last timestep, $\inv{\v\HH_{\cT}^{-}+\kap}\YY_{\cT}\rightarrow\inv{\v\HH^{\Tm\Tm}+\kap}\YY^{TT}$.

\paragraph*{Temporally-incoherent regime}

For small learning signal $\lambda$, the representation kernels $\HH$
of DNNs and RNNs coincide. The consistency of the diagonal solution
with the saddle point equations for the RNN can be seen as follows:
due to the masking operation in the DNN, a diagonal kernel is a stationary
point. Inserting such a diagonal solution into the saddle point equations
for the RNN, there are no terms that would cause off-diagonal correlations.
The diagonal solution hence obeys the saddle point conditions of the
RNN, too. What is unclear, though, is whether this is the only stationary
point. The RNN could have additional solutions with non-vanishing
off-diagonal elements that attain a higher likelihood than the diagonal
solution. Indeed for larger $\lambda$ this will be the case.\\
Since the output $f$ after learning depends only on the kernel $\HH$,
in the temporally-incoherent regime, both DNNs and RNNs yield the
same predictions and thus have the same generalization properties.

\paragraph{Temporally-coherent regime}

An explicit analytical form for the kernel $\HH$ can be derived for
the minimal case of $T=4$ (details in \prettyref{subsec:Landau_L4}).
As we show there, the continuous change in the order parameter is
the signature of a second-order phase transition with a critical exponent
$\hlf$, where the kernel's off-diagonal $\HH^{{\scriptscriptstyle T\shortmin1},{\scriptscriptstyle T\shortmin2}}$
forms the order parameter controlled by $\lambda$. \prettyref{fig:Landau_L4}
shows this transition for $T=4$ and for larger $T$. For the latter
case, to the best of our knowledge, no explicit form is available.
The transition prescribed by our mean-field theory is closely followed
by the kernel measured from direct SGLD experiments on the weights
$\bU,\bW,\bV$ (solid curves and diamonds, respectively, in \prettyref{fig:Landau_L4}\textbf{a}).
The phase transition thus casts learning as an optimization problem,
but with the weight loss $\L(\Theta;y,\bx)$ replaced by a lower-dimensional
objective over kernels, $-\ln P(\HH|y,\bx)/N$, which is shown in
\prettyref{fig:Landau_L4}.\textbf{ }Depending on the control parameter
$\lambda$, this kernel objective has a unique minimum only at the
temporally-incoherent solution when $\lambda$ is small, but this
minimum becomes unstable in favor of a new minimum at the coherent
solution beyond the transition point (see Fig. \ref{fig:Landau_L4}\textbf{b}).
\\
The analytics for $T=4$ developed in \prettyref{fig:Landau_L4} for
\prettyref{eq:P_H_lin} suggest an energy-entropy tradeoff underlying
this transition. In both architectures, the label likelihood in the
first line increases the final variance $\HH^{\Tm\Tm}$ by tying it
to $\YY^{{\scriptscriptstyle TT}}$ (``energy'', cf. \prettyref{eq:recurrent_net}).
However, this incurs a penalty on $\HH^{-}$ in $D_{\mathrm{KL}}\!\big(\mathcal{N}_{h}(\HH)\,\|\,\mathcal{N}_{h}(\w\mdiag{\HH^{-}}+\u\XX^{-})\big)$
that is the second line of \prettyref{eq:P_H_lin}, which encourages
the kernels to follow the NNGP iteration over time. In the DNN case,
the masking $\mdiag{\HH^{-}}=\text{diag}(\HH^{-})$ discards off-diagonals,
so introducing temporal correlations cannot reduce the $D_{\mathrm{KL}}$.
Thus, they remain suppressed by the $\lndet{\HH}$ term (an ``entropic''
penalty, since it favors uncorrelated $\HH$). In the RNN however,
the increase in $\HH^{\Tm\Tm}$ can be compensated by $\mdiag{\HH^{-}}=\HH^{-}$
through the introduction of temporal correlations. As it turns out,
the marginal cost of the entropic penalty from off-diagonals eventually
drops below the $D_{\mathrm{KL}}$ cost, once the labels $\YY$ are
sufficiently large relative to the priors $u,w,v$ that enter it.
Thus, eventually a transition point controlled by their ratio $\lambda$
is reached, separating the phases.

\subsection{RNN's weight sharing inductive bias generalizes efficiently for sequential
tasks}\label{subsec:efficient-seq-learners}

\begin{figure}
\centering
\centering{}\includegraphics[width=1\columnwidth]{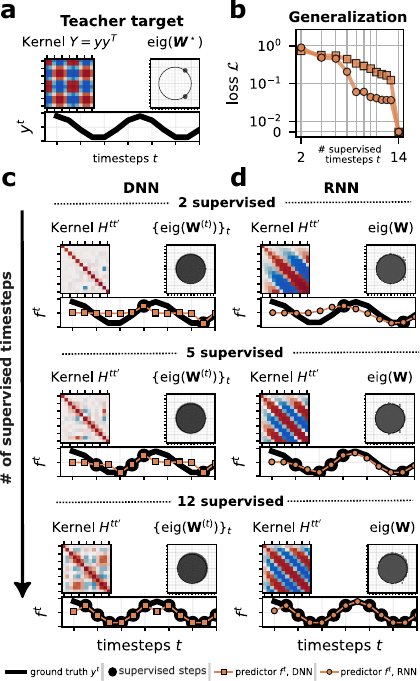}\caption{\textbf{RNNs have better sample efficiency in sequential tasks due
to task-model alignment induced by weight sharing.} \textbf{a)} Sequence
regression task, with supervision signal at variable numbers of timesteps
$t$. Top left: Label kernel $\protect\YY$. Top right: Spectrum of
the teacher weights $\protect\bW^{\star}$. Bottom: Sinusoidal output
of the teacher. \textbf{b)} Generalization error $\protect\L=\frac{1}{2T}\sum_{t=2}^{T}(y^{t}-f^{t})^{2}$
over all timesteps as a function of the number of supervised timesteps.\textbf{
c-d)} Representation $\protect\HH$ and output $f^{t}$ underlying
differences in generalization. DNN (\textbf{c}) and RNN (\textbf{d})
output for different number of supervision steps (rows). Each cell
displays the time-by-time kernel for pattern $p$ (top left), the
weight eigenspectra ($\{\protect\bW^{(t)}\}_{t}$ for DNN, $\protect\bW$
for RNN; top right), and on the bottom a comparison of the target
function $y^{t}$ (black line) versus the network prediction $f^{t}$
(orange) across supervised timepoints (black markers).}\label{fig:RNNs_are_efficient_sequence_learners}
\end{figure}
In \prettyref{subsec:RNN-DNN-comp}, we considered endpoint-supervised
tasks, identifying qualitative difference in the representation across
timesteps. While the last timestep's representation $\HH^{\Tm\Tm}$
exhibited quantitative differences in terms of its scale, its qualitative
structure was preserved, entailing similar predictors $f$. Since
such tasks do not read out from different timesteps and hence are
unaffected by temporal correlations, this leaves the question how
RNNs and DNNs differ for tasks with non-trivial temporal structure.
To investigate this, we now consider sequential tasks $x^{t}\mapsto y^{t}$,
where intuitively the RNNs' weight sharing may benefit generalization.\\
We consider a minimal setting where we have access to the ground truth
learning signal $y^{t}=x_{0}^{0}\sin(\omega t)+x_{1}^{0}\cos(\omega t)$
for $\bx^{0}=[\cos(\varphi_{0}),\,\sin(\varphi_{0})]^{\T}\in\bR^{D=2}$,
with a single input pattern $(P=1$). Notably, this task is identical
to a teacher RNN via $y^{t}=\bV^{\star}(\bW^{\star})^{t}\bU^{\star}\bx^{0}$,
with aligned readin $\bU^{\star}=\left[\begin{smallmatrix}1 & 0\\
0 & 1
\end{smallmatrix}\right]$, readout $\bV^{\star}=\left[\begin{smallmatrix}1\\
0
\end{smallmatrix}\right]$, and recurrent $\bW^{\star}=\left[\begin{smallmatrix}\cos(\Delta\varphi) & \:-\sin(\Delta\varphi)\\
\sin(\Delta\varphi) & \:\phantom{-}\cos(\Delta\varphi)
\end{smallmatrix}\right]$ weights. The teacher RNN dynamics thus describe a rotation of the
input vector $\bx^{0}$ in a 2D plane by an angle $\Delta\varphi$
in each timestep. In \prettyref{fig:RNNs_are_efficient_sequence_learners},
we find that RNNs trained on this task learn structured sequences
from fewer samples (since $P=1$, the number of supervised timesteps
takes the role of a sample count). From a Bayesian perspective, the
reason for this difference is the learned change in the kernel due
to the masking operation $\mdiag{\ldots}$ in \prettyref{eq:P_y__x},
which in turn determines the predictor via $f^{t}=\sum_{t't''}\HH^{tt'}(\inv{\HH+\kap})^{t't''}y^{t''}$
, where summations now go over timesteps. In the DNN, the posterior
covariance $\text{diag}(\HH)$ will always produce an incoherent representation
$h^{t}$ outside of the supervised training points, even after learning.
This means that generalization effectively becomes Bayesian inference
with an uncorrelated kernel, and thus extrapolating with the prior
mean, $f^{t}=0$. In contrast, the RNN's representation can develop
correlations in $\bh^{t}$ that match the target kernel $\YY$ already
after a few samples. In contrast to the DNN, this is accompanied by
the recovery of the conjugate pair of teachers in the eigenvalues
of $\bW$, pointing to a structured change in the spectrum. This difference
in generalization is perhaps surprising, since due to its weight sharing,
the RNN's expressivity is a subset of that of a DNN. \\
While this argument showcases that temporal correlations in $\HH$
can develop, it remains unclear what is the mechanism that shapes
them such that they indeed facilitate generalization. To understand
this, we consider the kernel theory $P(\HH|\YY,\XX)$ in \prettyref{eq:P_H_lin}
for the linear case and pursue a perturbative expansion in the label
strength $\YY$. Starting from the closed-form saddle-point equation
for the learned kernels (see \prettyref{eq:seq-perturb}, \prettyref{eq:MAP-closed-form-lin}),
we expand around the diagonal NNGP kernel $\HH_{0}$ (the solution
in absence of feature learning) by writing $\HH=\HH_{0}+\DD$, with
$\DD=\DD_{1}+\DD_{2}+\O(\YY^{3})$. The linear response $\DD_{1}$
turns out to not involve strong architectural effects, since masking
$\mdiag{\HH_{0}}$ is an identity operation on diagonal kernels. We
consider then the second order $\DD_{2}$ to identify architecture-dependent
effects. Introducing the NNGP observation covariance $\GG_{y}\coloneqq\inv{\v\HH_{0}+\kap}$,
the leading correction can be written as 
\begin{align}
 & \HH-\HH_{0}\approx+\GG_{y}\YY^{\smlplus}\GG_{y} & \Bigr\}{\scriptstyle =\Delta_{1}}\label{eq:seq-perturb}\\
 & -\bigl(\w^{{\scriptscriptstyle \shortmin2}}\,\rinv{\HH_{0}}\mdiag{\DD_{1}}\,\rinv{\HH_{0}}\,\DD_{1}\rinv{\HH_{0}}\,+\,(\sim)^{\T}\bigr)\,+\,\dldots & \Bigr\}{\scriptstyle =\Delta_{2}}\nonumber 
\end{align}
where $\mdiag{\ldots}$ again is the identity for RNNs and a diagonal
projection for DNNs, $\YY^{\smlplus}$ denotes time shift, $(\sim)^{\T}$
is the transpose of preceding term, and we omitted terms independent
of masking ``$\dldots$'' or that are of higher order $\O(\YY^{3})$.\\
To interpret this expression, consider a time-diagonal $\XX$ and
a kernel $\HH_{0}$ \citep{Segadlo22_accepted} with \textit{constant}
diagonal, conditions which are both met for \prettyref{fig:RNNs_are_efficient_sequence_learners}
for $u=w=v=1$. Then, $\HH_{0}^{\smlplus}$ is diagonal with entries
$\mathds{h}_{0}^{t}$ and we may write $\DD_{1}=\GG_{y}\YY^{\smlplus}\GG_{y}\propto\YY^{\smlplus}$.
\\
In a DNN, $\mdiag{\DD_{1}}=\text{diag}(\DD_{1})$ and the quadratic
term acts entry-wise, 
\begin{equation}
\DD_{2,\text{DNN}}^{tt'}=-\finv{\mathds{h}_{0}^{t}}\bigl(\finv{\mathds{h}_{0}^{t}}\YY^{\smlplus tt}+\finv{\mathds{h}_{0}^{t'}}\YY^{\smlplus t't'}\bigr)\YY^{\smlplus tt'}\finv{\mathds{h}_{0}^{t'}}+\dldots,\label{eq:taylor-DNN}
\end{equation}
In an RNN, in contrast, we get
\begin{equation}
\DD_{2,\text{RNN}}^{tt'}=-2\,\finv{\mathds{h}_{0}^{t'}}\bigl(\sum_{t''}\YY^{\smlplus tt''}\finv{\mathds{h}_{0}^{t''}}\YY^{\smlplus t''t'}\bigr)\finv{\mathds{h}_{0}^{t'}}+\dldots\,.\label{eq:taylor-RNN}
\end{equation}
This equation shows that the inverses appearing in \prettyref{eq:seq-perturb}
take the role of \textit{propagators}, and hence \prettyref{eq:taylor-RNN}
implements an interpolation between supervised points of $\YY$. To
see this explicitly, consider the case that $t_{1}$ and $t_{2}$
have been observed but $t_{3}$ has not. Then, the correlation $\DD_{2,\text{RNN}}^{t_{3}t_{1}}$
will interpolate across the path $t_{3}\,\leftarrow\,t_{2}\,\leftarrow\,t_{1}$
that appears in the sum at $t'=t_{2}$, but $\DD_{2,\text{DNN}}^{t_{3}t_{1}}=0$
in \prettyref{eq:taylor-DNN}, making the posterior over $h$ uncorrelated.\\
A complementary way to understand this is that $\DD_{1}$ is an $\HH_{0}$-whitened
version of $\YY^{\smlplus}$, and due to $\HH_{0}$'s diagonal structure
will approximately inherit its eigenvectors. For the DNN, the masking
in the second-order iteration \prettyref{eq:taylor-DNN} will destroy
this structure, so that learning requires more samples.\\
In summary, this shows that the mechanism behind the inductive bias
of weight sharing can be traced back to the propagation of label messages
across unsupervised time points. 

\section{Discussion}

\textbf{Summary.} In this work, we developed a unified theory of feature
learning for RNNs and DNNs in terms of representational similarity
kernels $\Pfi=\fN\sum_{i}^{N}\phi(h_{i})\phi(h_{i})^{\T}$ and $\HH=\fN\sum_{i}^{N}h_{i}h_{i}^{\T}$,
describing trained networks in the feature learning ($\mu$P) regime.
This theory reveals the functional implications of architectural structure:
it describes how the representational kernels $\Pfi$ and $\HH$ as
well as the predictor $f$ are shaped by weight scales $u,w,v$, the
structure of input $\XX$ and labels $\YY$, and the network architecture
itself. In this theory, the key architectural difference between RNNs
and DNNs, weight sharing across timesteps (layers), becomes a masking
operation $\mdiag{\Pfi}$ on the temporal correlations of hidden activities
$\phi(h)$ in Equation \prettyref{eq:P_y__x}, affecting the activities
$h$ and the predictor $f$.

\textbf{Phases of feature learning.} Feature learning is often characterized
in terms of $\O(1)$ changes in the representation~\citep{Yang20_14522,bordelon2022}.
We find that, within this regime, a sufficiently strong learning signal
$\lambda$ is necessary in addition to drive \textit{structured} alignment
in RNNs: not merely changes in eigenvalues of the representational
kernel, but alignment of its eigenvectors to the task covariance.
This suggests a more granular view of feature learning beyond uniform
changes in scale~\citep{canatar2021,Li21_031059}: the learned eigenvectors
themselves constitute the features that improve generalization, consistent
with how the term ``feature'' is used in neuroscience~\citep{Hubel62,Yamins13}.
The phase transition we identify is reminiscent of a Baik-Ben~Arous-Péché
(BBP) transition~\citep{baik2004}, where random initialization acts
as noise that gradient signals must overcome, mirroring similar findings
by \citet{bordelon2025}.

\textbf{Connection to initialization.} In practice, feature learning
is often controlled via initialization scale, small initialization
leading to stronger representation of data features in the network
learned. In our theory, the parameters $u,w,v$ control a Gaussian
prior over weights, and a narrow prior (large $\lambda=\nicefrac{y^{2}}{\u\w^{T-2}\v}$)
drives the representation from \textit{temporally incoherent} to \textit{temporally
coherent}. These two perspectives match in the small noise regime
$\kappa\ll1$, where the diffusion over the Gaussian prior becomes
much slower than changes due to the loss. Our theory thus details
why small initialization aids feature learning.

\textbf{Inductive bias and task-model alignment.} A central finding
is that weight sharing endows RNNs with an inductive bias that can
outweigh the DNN's greater expressivity. In sequential tasks, RNNs
generalize from fewer samples due to a representation which infers
the task structure, whereas DNNs effectively revert to a white prior
outside supervised timesteps. This is in line with ``No Free Lunch''
theorems~\citep{wolpert1996lack}: the larger parameter count of
DNNs need not be universally better when the inductive bias of a more
constrained model matches the task structure. Our theory provides
a mechanism behind this phenomenon in terms of interpolation between
labels.

\textbf{Limitations and future work.} Regarding scope, our theory
deliberately chooses a functional perspective by marginalizing over
the weights, thereby abstracting away their learning dynamics and
structure. Regarding training, we approximated the correlated mini-batch
noise in plain SGD with independent noise of SGLD \citep{welling2011bayesian}.
Regarding tasks, we considered minimal settings to identify foundational
mechanisms of learning via interpretable analytics; notably linear
networks and orthogonal inputs. This choice has allowed us to focus
on temporal generalization. However, there are important other phenomena
in deep learning that depend on the structure over patterns $p$ \citep{sclocchi2024}
or their interaction with timesteps $t$, particularly in proportional
limits like $T,P\propto N$. Likewise, representations will be shaped
by the cumulative application of nonlinearities across layers~\citep{Keup22_11355}.
While in principle our theory describes these cases as well, they
go beyond the scope of this work.

\clearpage{}

\subsection*{Acknowledgements}

We would like to thank Alexander van Meegen and Antonio Sclocchi for
helpful discussions. This work is supported by the Gatsby Charitable
Foundation (GAT3850 to JPB and AP), and the Simons Foundation (1156607
to AP) and by the Deutsche Forschungsgemeinschaft (DFG, German Research
Foundation) - 368482240/GRK2416, the Helmholtz Association Initiative
and Networking Fund under project number SO-092 (Advanced Computing
Architectures, ACA), the Deutsche Forschungsgemeinschaft (DFG, German
Research Foundation) as part of the SPP 2205 -- 533396241, and the
DFG grant 561027837/HE 9032/4-1 (to MH).

\subsection*{Impact statement}

This paper presents work whose goal is to advance the field of Machine
Learning. There are many potential societal consequences of our work,
none which we feel must be specifically highlighted here.

\newpage{}

\bibliographystyle{icml2026}\bibliography{brain,bib,cited}

\clearpage{}

\appendix

\onecolumn

\section{Appendix}

\counterwithin{figure}{section} %

\subsection*{Overview}\label{sec:appendix-overview}

We structure the Appendix as follows: we first in \prettyref{subsec:Bayes-LSGD}
establish the connection between Bayesian inference and the stationary
distribution of stochastic gradient Langevin dynamics (SGLD). In \prettyref{sec:Derivation-of-network-prior},
we derive the infinite-width network prior and the resulting saddle-point
equations for the network's kernel. We detail the effects of partial
supervision in \prettyref{subsec:Effect-of-partial} and the reduction
to the standard DNN architecture in \prettyref{subsec:Reduction-to-DNN}.
The section \prettyref{sec:linear} presents an analytical study of
linear recurrent networks, deriving closed-form kernel solutions and
performing a Landau analysis of the symmetry-breaking transition.
Finally, we outline the numerical solution methods and provide a step-by-step
summary of the derivation in \prettyref{subsec:Summary-mft-derivation}.

\subsection{Setup}\label{subsec:Bayes-LSGD}

Recall the definition of the RNN: at each time step $t$, the preactivations
are $\bh^{t}\in\mathbb{R}^{N\times P}$, where $P$ is the index over
training samples. The networks have one scalar output for each training
sample $y^{t}\in\mathbb{R}^{P}$ and are updated according to
\begin{align}
\bh^{t} & =\bW\bphi^{t-1}+\bU\bx^{t-1}\,\quad1\le t<T\,,\label{eq:recurrent_net}\\
f^{t+1} & =\bV\bphi^{t}\,,\nonumber \\
y^{t+1} & =f^{t+1}+\xi^{t+1}\,,\nonumber 
\end{align}
where $\bx^{t}$ is the input at time $t$ and we have defined the
shorthand $\bphi^{t}=\phi(\bh^{t})$. The parameters $\bU\in\mathbb{R}^{N\times D}$,
$\bW\in\mathbb{R}^{N\times N}$, $\bV\in\mathbb{R}^{1\times N}$ are
matrices which we assume to have Gaussian priors that are i.i.d. over
the matrices' entries

\begin{align}
\bU & \overset{\iid}{\sim}\N_{\bU}(\bU)\coloneqq\N(0,\,G_{\bU})\,,\label{eq:priors}\\
\bW & \overset{\text{\ensuremath{\iid}}}{\sim}\N_{\bW}(\bW)\coloneqq\N(0,\,G_{\bW})\,,\nonumber \\
\bV & \overset{\iid}{\sim}\N_{\bV}(\bV)\coloneqq\N(0,\,G_{\bV})\,,\nonumber 
\end{align}
for which our derivation will reveal the natural scalings $G_{\bU}\eqqcolon U=\nicefrac{\u}{D},\,G_{\bW}\eqqcolon W=\nicefrac{\w}{N},\,G_{\bV}\eqqcolon V=\nicefrac{\v}{N^{2}}$
with $\O(1)$ parameters $\v,\w,\u$. The network output $f^{t}$
is a vector $\in\mathbb{R}^{P}.$ Furthermore, we assume a readout
noise that is i.i.d. in time and patterns, $\xi_{p}^{t}\overset{\text{i.i.d.}}{\sim}\N_{\Kap}(\xi_{p}^{t})\coloneqq\N(0,\,\Kap)$
with $\Kap=\nicefrac{\kap}{N}$. We define time-advanced $(\circ^{+})^{tt'}\coloneqq\mathbb{\circ}^{t+1,t'+1}$
and time-retarded $(\mathbb{\circ}^{-})^{tt'}\coloneqq\mathbb{\circ}^{t-1,t'-1}$
kernel matrices to ease notation. We also define the input kernel
$\XX^{tt'}\coloneqq\frac{1}{D}\sum_{i=1}^{D}x_{i}^{t}x_{i}^{t'}$
and the label kernel $\YY^{tt'}\coloneqq y^{t}y^{t'}$ for convenience. 

\subsection{Bayesian inference and relation to training dynamics}\label{subsec:Bayesian-inference-and-LSGD}

\global\long\def\vf{f}%
\global\long\def\vy{y}%
\global\long\def\vx{\bx}%

We here derive a theory of learning that has two interpretations.
The first is that of Bayesian inference. Second, as the stationary
distribution of weights after training by stochastic gradient Langevin
dynamics (SGLD). Both approaches will be described by the same set
of equations.

\subparagraph{Bayesian inference}

Given a set of training data $\D=\{(\bx_{p}^{t-1},y_{p}^{t+1})\}_{1\le p\le P}^{1\le t<T}$,
the Bayesian approach assumes a neuronal architecture, for example
\eqref{eq:recurrent_net}, which defines the network output as a function
$f(\Theta,\bx)$ that depends on the network's input $\bx\in\bR^{\Tm\times P\times D}$
and its parameters $\Theta$; in our case, $\Theta=\{\bW,\bV,\bU\}$.
The set of all inputs is combined into the matrix $\bx$ and the set
of outputs in a vector $y\in\bR^{\Tm\times P}$. One assumes a prior
distribution $P(\Theta)$ on the set of parameters; in our case, the
distribution $P(\Theta)$ is given by the set of distributions $\N_{\bU}(\bU)$,
$\N_{\bW}(\bW)$, $\N_{\bV}(\bV)$ \eqref{eq:priors}.

Together with the observation likelihood $\N\big(\vy|\vf(\Theta,\vx),\Kap\,\bI\big)$,
this defines a posterior over weights
\begin{equation}
P(\Theta|y,\bx)\propto\N\big(\vy|\vf(\Theta,\vx),\Kap\,\bI\big)\,P(\Theta),\label{eq:weight_posterior}
\end{equation}
using the rules of Bayesian inference. In turn, this weight posterior
induces a marginal distribution on labels $y$
\begin{align}
P(\vy|\vx) & =\int P(y,\Theta|\bx)\,d\Theta.\label{eq:network_prior}
\end{align}

We can obtain the statistics of any function $O(\Theta)$ under the
posterior over weights from the cumulant generating function (sometimes
referred to as ``free energy'')
\begin{align}
\mathcal{W}(\vy,j) & =\ln\,\int e^{j\,O(\Theta)}\,\N\big(\vy|\vf(\Theta,\vx),\Kap\,\bI\big)\,P(\Theta)\,d\Theta\,.\label{eq:posterior_observable}
\end{align}
For example, the mean of $O$ follows as
\begin{align*}
\vev{O(\Theta)}_{P(\Theta|\bx,y)} & =\partial_{j}\,\mathcal{W}(\vy,j)\big|_{j=0}\,,
\end{align*}
where one notes that the outer derivative of the logarithm yields
the correct normalization as in \eqref{eq:weight_posterior}. Higher
order cumulants are obtained as higher order derivatives.

\subparagraph{Stochastic gradient Langevin dynamics}

The second interpretation of \eqref{eq:weight_posterior} is that
of a stationary distribution of a time-dependent learning rule for
the weights. It is known that the stochastic differential equation
\begin{align}
d\Theta(s) & =-\nabla_{\Theta}H(\Theta)\,ds+dB(s)\,,\label{eq:Langevin_training}\\
\langle dB(s)dB(s')\rangle & =\delta(s-s')\,\frac{2}{\beta}\,ds\,,\nonumber 
\end{align}
where $dB(t)$ is a Wiener increment, has the stationary distribution
(e.g., \citet{Gardiner85,Risken96})
\begin{align}
P(\Theta) & \propto\exp\big(-\beta\,H(\Theta)\big)\,.\label{eq:Boltzmann}
\end{align}
For Gaussian prior measures on the weights with variance $g$ one
has $P(\Theta)\propto\exp\big(-\|\Theta\|^{2}/{\scriptstyle 2G}\,\big)$,
so that \eqref{eq:network_prior} can be written as
\begin{align*}
P(\vy|\vx) & \propto\int\exp\big(-\frac{1}{2\Kap}\|\vy-\vf(\Theta,\vx)\|^{2}-\frac{1}{2G}\|\Theta\|^{2}\big)\,d\Theta\,,
\end{align*}
so that we identify $H$ in \eqref{eq:Boltzmann} as 
\begin{align*}
H(\Theta) & =\frac{P\Tm}{\Kap\beta}\,\L(\Theta)+\frac{1}{2G\beta}\|\Theta\|^{2}\,,
\end{align*}
where the first term is expressed in terms of the the mean squared
error between the given data and network output

\[
\L(\Theta)=\frac{1}{2P\Tm}\|\vy-\vf(\Theta,\vx)\|{}^{2}\,,
\]
where the appearing vectors in the norm are $\mathbb{R}^{\Tm P}$

Discretizing \prettyref{eq:Langevin_training} in time and omitting
a Metropolis-Hastings correction (MALA) \citet{besag1994comments}
, this scheme becomes identical to plain stochastic gradient descent
(SGD), with the difference that the source of the noise here is $\iid$
Gaussian, whereas it is more structured mini-batch noise in typical
deployments of SGD.

\subsection{Bayesian adaptive kernel theory for RNN and DNN}\label{subsec:MFT}

In this section we derive a unified mean-field formulation in kernel-space
that treats deep feedforward (DNN) and recurrent (RNN) networks on
the same footing. This is achieved by unrolling the RNN in time. The
strategy is to treat learning as Bayesian inference and to write the
network as a conditional density $P(f|\bx)$ that is an integral over
collective fields encoding pre-activation covariances, integrate out
all microscopic Gaussian parameters, and then take the large-width
limit $N\to\infty$ via a saddle-point analysis. In this formulation
the only structural difference between DNNs and RNNs is encoded by
a simple masking operator on the time indices (equivalent to layers
in the DNN case), while the scaling of the parameters and auxiliary
fields with networks width $N$ remains identical across architectures.

\subsubsection{Derivation of network prior}\label{sec:Derivation-of-network-prior}

We here derive the concrete form of the network prior \eqref{eq:network_prior}
for the architecture \eqref{eq:recurrent_net}. The computation follows
closely previous works \citet{Segadlo22_accepted,Fischer24_10761,lauditi2025}.
We also give a compact overview of the derivation in \prettyref{subsec:Summary-mft-derivation}
that emphasizes the choice of scaling.

Since the cumulants of an observable \eqref{eq:posterior_observable}
over the posterior over weights can be obtained equivalently from
the network prior \eqref{eq:network_prior} by taking the logarithm
and considering suitable derivatives, it is sufficient to derive the
form of the prior. To this end, it is convenient to decompose the
network prior into a chain of conditional probabilities
\begin{align}
P(\vy|\vx) & =\int\N\big(\vy|\vf',\Kap\,\bI\big)\,P(\vf'|\vx)\,d\vf'\,,\label{eq:decompose_as_chain_of_prob1}
\end{align}
with 
\begin{align*}
P(\vf'|\vx) & =\vev{\delta(\vf'-\vf(\bU,\bV,\bW,\vx))}_{\bU\sim\N_{\bU}\,,\bV\sim\N_{\bV}\,,\bW\sim\N_{\bW}}.
\end{align*}
The latter can further be decomposed as
\begin{align}
P(\vf'|\vx) & =\int P(\vf'|\bh)\,P(\bh|\vx)\,d{\bf h}\,.\label{eq:decompose_as_chain_of_prob2}
\end{align}
The Dirac constraint is rewritten in its Fourier representation as
\begin{align*}
\delta(\vf'-\vf) & \equiv\prod_{p=1}^{P}\prod_{t=2}^{T}\delta(f_{p}^{\prime t}-f_{p}^{t})\\
 & =\prod_{p=1}^{P}\prod_{t=2}^{T}\,\frac{1}{2\pi i}\int_{-i\infty}^{i\infty}d\tf_{p}^{t}\,\exp\Bigl\{\tf_{p}^{t}\,\big[f_{p}^{\prime t}-f_{p}^{t}\big]\Bigr\}\,,
\end{align*}
which allows us to compute the expectation value over $\bV$ as
\begin{align*}
P(f^{\prime}|\bh)=\int\mathcal{D}\tf\, & \Big\langle\exp\Bigl\{\sum_{t,p}\,\tf_{p}^{t}\,\big[f_{p}^{\prime t}-\sum_{i}V_{i}\phi_{p,i}^{t-1}\big]\Bigr\}\,\Big\rangle_{\bV\sim\N_{\bV}(0,\nicefrac{\v}{N^{2}})}\\
= & \exp\Bigl\{\sum_{t,p}\,\tf_{p}^{t}\,f_{p}^{\prime t}+\frac{1}{2}\sum_{p,p';t,t'}\,\tf_{p}^{t}\tf_{p'}^{t'}\,\frac{v}{N^{2}}\sum_{i}\phi_{p,i}^{t-1}\phi_{p',i}^{t'-1}\Bigr\}\,,
\end{align*}
where we define the integral measure $\int\mathcal{D}\tf=\Big(\prod_{p=1}^{P}\prod_{t=2}^{T}\,\frac{1}{2\pi i}\int_{-i\infty}^{i\infty}d\tf_{p}^{t}\Big)$.
We write the latter expression for short as
\begin{align}
P(f^{\prime}|\bh)=\int\mathcal{D}\tf\,\exp\Bigl\{\sum_{t}\tf^{t\T}f^{\prime t}+\frac{1}{2}\sum_{t,t'}\,\tf^{t\T}\frac{v}{N^{2}}\big[\bphi{}^{t-1}\cdot\bphi^{t'-1\T}\big]\,\tf^{t'}\Bigr\}\,,\label{eq:disorder_V}
\end{align}
where $\bm{a}\cdot\bm{b}$ denotes an inner product over $i=1,\ldots,N$
and $a^{\T}b$ an inner product over $p=1,\ldots,P$. We showed how
we can write \eqref{eq:decompose_as_chain_of_prob1} as:

\begin{align}
P(\vy|\vx) & =\int\Big[\int\N\big(\vy|\vf',\Kap\,\bI\big)\,P(\vf'|\bh)\,d\vf'\,,\Big]P(\bh|\vx)\,d\bh\label{eq:P_y_x_expanded}
\end{align}

Performing the integral over $f'$, yields 
\begin{align}
P(\vy|\bh) & =\int\N\big(y|f',\Kap\,\bI\big)\,P(\vf^{\prime}|\bh)\,d\vf^{\prime}\label{eq:p_y_h_integrated}\\
 & =\int\mathcal{D}\tf\,\exp\Bigl\{\sum_{t}\tf^{t\T}\vy^{t}+\frac{1}{2}\sum_{t}\,\frac{\kappa}{N}\tf^{t\T}\tf^{t}+\frac{1}{2}\sum_{t,t'}\,\tf^{t\T}\frac{v}{N^{2}}\big[\bphi{}^{t-1}\cdot\bphi^{t'-1\T}\big]\,\tf^{t'}\Bigr\}\,,\nonumber 
\end{align}
where we used the   parameter $\Kap=\kappa/N$. We again use the
Fourier representation of the delta function to enforce the evolution
of the network dynamics \eqref{eq:recurrent_net} for each sample
$p$ and each time $t$ as
\begin{align}
P(\bh|\vx)= & \prod_{t}\,\vev{\delta(\bh^{t}-\bW^{(t-1)}\bphi^{t-1}-\bU\bx^{t-1})}{}_{\bU\sim\N_{\bU},\,\{\bW^{(t)}\}\sim\N_{\bW}}\label{eq:disorder_W_U}\\
= & \int\mathcal{D}\tilde{\bh}\,\exp\Bigl\{\sum_{t}\,\tilde{\bh}^{t\T}\cdot\bh^{t}+\frac{1}{2}\sum_{t,t'}\,\tilde{\bh}^{t}\cdot\Big[\bar{\delta}^{t-1,t'-1}\frac{w}{N}\,\bphi^{t-1}\cdot\bphi^{\T t'-1}+\frac{u}{D}\,\bx^{t-1}\cdot\bx^{t'-1\T}\big]\cdot\tilde{\bh}^{t'}\Bigr\}\,.\nonumber 
\end{align}
Where we used the identity $\mathbb{E}_{w\sim\mathcal{N}(0,\Sigma)}\!\left[e^{a^{\top}w}\right]=\exp\!\left(\frac{1}{2}\,a^{\top}\Sigma\,a\right)$
and introduced the symbol
\begin{align*}
\bar{\delta}^{tt'}= & \begin{cases}
\delta^{tt'} & \text{DNN}\\
1 & \text{RNN}
\end{cases}\,.
\end{align*}
The difference arises, because for the RNN there is only a single
weight matrix $\bW$ valid for all timesteps $t$, thus its entries
are perfectly correlated across time, while in the DNN, the matrices
$\bW^{(t)}$ are drawn from prior distributions which are independent
across different layers $t$.

One notes that due to the appearance of the inner products$\cdot$
of the fields $\tilde{\bh}$, the exponent factorizes across neuron
indices $i$, which allows us to reduce the $N$ integrals over $\bh$
and $\tilde{\bh}$ to a single integral each. We introduce auxiliary
fields for terms that contain an inner product over the neuron index.
which we anticipate will concentrate in the large-$N$ limit, thus
defining
\begin{align}
\Pfi^{tt'} & :=\frac{1}{N}\,\bphi^{t}\cdot\bphi^{t'\T}\in\mathbb{R}^{P\times P}\,,\label{eq:def_aux_fields}\\
\XX^{tt'} & :=\frac{1}{D}\,\bx^{t}\cdot\bx^{t'\T}\in\mathbb{R}^{P\times P}\,.\nonumber 
\end{align}
Likewise, the kernels $\XX_{pp'}^{tt'}\coloneqq\frac{1}{D}\sum_{i}^{D}x_{p,i}^{t}x_{p',i}^{t'}$
and $\YY_{pp'}^{tt'}\coloneqq y_{p}^{t}y_{p'}^{t'}$ denote input
and label kernels, and $\HH_{pp'}^{tt'}\coloneqq\fN\sum_{i}^{N}h_{p,i}^{t}h_{p',i}^{t'}$
is the kernel for the preactivations, defined in analogy to the kernel
$\Pfi$ for the postactivations. Enforcing the former definition by
introducing a second auxiliary field $\dbltilde C\in i\,\mathbb{R}^{PT\times PT}$
one inserts a Dirac constraint in the form 
\begin{align*}
\delta(-\Pfi^{tt'}+\fN\bphi^{t}\cdot\bphi^{t'\T}) & =\int\D\dbltilde C\,\exp\Bigl\{-\sum_{t,t';p,p'}\dbltilde C_{pp'}^{tt'}\Pfi_{pp'}^{tt'}+\dbltilde C_{pp'}^{t,t'}\,\fN\bphi_{p}^{t}\cdot\bphi_{p'}^{t'}\Bigr\}\,.
\end{align*}

where we used $\int\mathcal{D}\tilde{\tilde{C}}=\Big(\prod_{p,p'=1}^{P}\prod_{t,t'=1}^{\Tm}\,\frac{1}{2\pi i}\int_{-i\infty}^{i\infty}d\tilde{\tilde{C_{pp'}^{tt'}\big)}}$.
This allows us to use equation \eqref{eq:disorder_V} to write 
as
\begin{align*}
P(\vy|\Pfi,\vx) & =\int\mathcal{D}\vf\,\exp\Bigl\{\vf^{\T}y+\frac{1}{2}\sum_{t,t'}\,\vf^{t\T}(\frac{v}{N}\,\Pfi^{t-1,t'-1}+\frac{\kappa}{N})\,\vf^{t'}\Bigr\}\,,\\
 & =\exp\Bigl\{-\frac{N}{2}\sum_{tt'}y^{t\T}(\inv{\v\Pfi^{-}+\kappa})^{tt'}y^{t'}-\frac{1}{2}\ln|\v\Pfi^{-}+\kappa|+\frac{P\Tm}{2}\ln N\Bigr\}\,,
\end{align*}
where we performed the Gaussian integral over $\tf$ in \eqref{eq:p_y_h_integrated}
and introduced the shorthand $\Pfi^{-}$ as the one time-step shifted
version of the original matrix, i.e. $(\circ^{-})^{tt'}\coloneqq\circ^{t-1,t'-1}$.
Inserting \eqref{eq:disorder_W_U} and \eqref{eq:disorder_W_U} in
\eqref{eq:P_y_x_expanded}, and inserting the definitions of the auxiliary
fields \eqref{eq:def_aux_fields} we obtain .

\begin{align}
P(y|\vx) & \propto\int\D\Pfi\,\exp\left\{ -\frac{N}{2}\sum_{tt'}y^{t\T}(\inv{\v\Pfi^{-}+\kappa})^{tt'}y^{t'}+\order(1)\right\} \label{eq:p_y_X_final}\\
 & \times\int\D\dbltilde C\,\exp\Bigl\{-\sum_{tt'}\dbltilde C^{tt'\T}\Pfi^{tt'}+N\,\mathcal{W}(\dbltilde C/N|\text{\ensuremath{\Pfi},}\XX)\Bigr\}\,,\nonumber \\
\nonumber \\\cW(\tPfi|\text{\ensuremath{\Pfi},}\XX) & :=\ln\,\int\mathcal{D}\th\,\exp\Bigl\{\sum_{t,t'}\phi^{t\T}\tPfi^{t,t'}\,\phi^{t'}+\sum_{t}\,\th^{t\T}\th^{t}+\frac{1}{2}\sum_{t,t'}\,\th^{t\T}\Big[\w\bar{\delta}^{tt'}\ensuremath{\Pfi}^{t-1,t'-1}+\u\XX^{t-1,t'-1}\big]\,\th^{t'}\Bigr\}\,,\nonumber 
\end{align}
where we used that the integrals over $\bh$ and over $\tilde{\bh}$
factorize over neurons $i=1\ldots N$ and thus yield the same integral
to the $N$-th power -- hence the factor $N\,\mathcal{W}$ appearing
and we dropped the term $\frac{1}{2}\ln|\v\ensuremath{\Pfi}^{-}+\kappa|$
in the first line that scales as order unity on $N$ as well as constant
terms. A short way of writing the latter line is as
\begin{align}
\cW(\tPfi|\Pfi,\XX) & :=\ln\,\Bigg\langle\,\exp\Bigl\{\sum_{t,t'}\sum_{p,p'}\tPfi_{pp'}^{tt'}\,\phi(h_{p}^{t})\phi(h_{p'}^{t'})\Bigr\}\,\Bigg\rangle_{h\sim\N(0,\,\w\bar{\delta}^{tt'}\Pfi^{t-1,t'-1}+\u\XX^{t-1,t'-1})}\,,\label{eq:def_calW}
\end{align}
which also corresponds to taking the integral over the $\th$ fields.

We note that $N\,\cW(\dbltilde C/N|\ensuremath{\Pfi},\XX)$ has the
form of a cumulant-generating function for the random variable $\Pfi$
and hence $P(\Pfi|\XX)=\int\D\dbltilde C\,\exp\big(-\sum_{tt'}\dbltilde C^{tt'\T}\Pfi^{tt'}+N\,\cW(\dbltilde C/N|\ensuremath{\Pfi},\XX)\big)$
is the Fourier representation of the probability distribution $P(\Pfi|\XX)$.
The trailing factor $N$ and the factor $N^{-1}$ coming together
with the source variable come in what is known as the scaling form
-- they indicate that first derivatives are $\order(1)$, while all
higher derivatives are suppressed with at least $N^{-1}$, indicating
that the mean dominates the distribution of $\Pfi$.

An alternative way of seeing this is to define the rescaled, intensive
field $\tPfi:=\dbltilde C/N$ which then leads to
\begin{align}
P(\ensuremath{\Pfi}|\XX) & \propto\int\D\tPfi\,\exp\Bigl\{-N\,\sum_{tt'}\tPfi^{tt'\T}\Pfi^{tt'}+N\,\cW(\tPfi|\ensuremath{\Pfi},\XX)\Bigr\}\nonumber \\
 & \stackrel{N\to\infty}{\simeq}\exp\Bigl\{ N\,\sup_{\tPfi}\left[-\sum_{tt'}\tPfi^{tt'\T}\Pfi^{tt'}+\cW(\tPfi|\ensuremath{\Pfi},\XX)\right]\Bigr\}\nonumber \\
 & =:\exp\Bigl\{-N\,\Gamma(\ensuremath{\Pfi}|\XX)\,\Bigr\}.\label{eq:def_gamma}
\end{align}
The latter expression follows from a saddle point approximation of
the Fourier integral over $\tPfi$ and shows that a rate function
$\Gamma$ appears in the exponent. We note that the entire integrand
appearing in \eqref{eq:p_y_X_final} has a trailing factor $N$, so
that we may also take the integral over $\ensuremath{\Pfi}$ in saddle
point approximation, which leads to
\begin{align}
\ln\,P(\vy|\vx)/N & \simeq\sup_{\ensuremath{\Pfi}}\,\Big[-\frac{1}{2}\sum_{tt'}y^{t\T}(\inv{\v\Pfi^{-}+\kappa})^{tt'}y^{t'}-\Gamma(\Pfi|\XX)\Big]=:S(\ensuremath{\Pfi}|\YY,\XX)\,.\label{eq:def_S_phi}
\end{align}
This object is our main theoretical result, which recasts the problem
of learning into a variational problem of determining the maximum
of the right hand side with regard to $\Pfi$.
\subsubsection{Saddle-point equations for the RNN case}

We now instantiate the RNN case by replacing $\bar{\delta}^{tt'}=1$
and redefine contractions $a^{\T}b\coloneqq\sum_{tp}a_{p}^{t}b_{p}^{t}$
to run over patters and time  (unlike only patterns as before) to
ease notation. The supremum condition (saddle point equation) for
$\tPfi$ can be obtained by taking in \eqref{eq:def_gamma} then becomes

\[
\begin{array}{cc}
\Pfi_{pp'}^{tt'}=\frac{\partial\cW}{\partial\tPfi_{pp'}^{tt'}}(\tPfi|\YY,\XX)= & \frac{1}{Z}\vev{\phi(h_{p}^{t})\phi(h_{p'}^{t'})\,\exp\{\sum_{pp'}\sum_{tt'}\,\phi(h_{p}^{t})\tPfi_{pp'}^{tt'}\phi(h_{p'}^{t'})\}}_{h\sim\N(0,\,w\Pfi^{-}+u\XX^{-})}\\
 & \eqqcolon\vev{\phi(h_{p}^{t})\phi(h_{p'}^{t'})}_{P(h|y,\bx)},
\end{array}
\]
where $Z=\vev{\exp\{\sum_{pp'}\sum_{tt'}\,\phi(h_{p}^{t})\tPfi_{pp'}^{tt'}\phi(h_{p'}^{t'})\}}_{h\sim\N(0,\:w\Pfi^{-}+u\XX^{-})}$
is a the normalization constant. The expectation thus is taken with
respect to the posterior $P(h|y,\bx)=P(h|\Pfi,\tPfi,\YY,\XX)$ , i.e.
the kernels $\Pfi,\tPfi,\YY,\XX$ are sufficient statistics for this
measure. 

To close these equations, we still need an expression for $\tPfi$.
For that we re-write the action $S(\ensuremath{\Pfi}|\YY,\XX)=\sup_{\ensuremath{\Pfi}}\,\Big[-\frac{1}{2}\tr\Big(\YY\inv{\v\Pfi^{-}+\kappa})\Big)-\Gamma(\Pfi|\XX)\Big]$
and the cumulant generating function \eqref{eq:def_calW} 
\begin{align*}
\cW(\tPfi|\XX) & :=\ln\,\int dh\,\,\exp\Big(-\frac{1}{2}h^{\T}\inv{\w\mdiag{\Pfi^{-}}+\u\XX^{-}}h+\phi^{\T}\tPfi\,\phi\Big)-\frac{1}{2}\ln|\w\mdiag{\Pfi^{-}}+\u\XX^{-}|\,,
\end{align*}
in term of kernel matrices. We had introduced the kernel $\HH_{pp'}^{tt'}\coloneqq\fN\sum_{i}^{N}h_{p,i}^{t}h_{p',i}^{t'}$,
which we can now write as $\HH\coloneqq\vev{hh^{\T}}_{P(h|\vx,y)}$.
Using $\frac{\partial}{\partial\Pfi^{-}}\,\ln|\w\mdiag{\Pfi^{-}}+\u\XX^{-}|=\w\,\inv{\w\mdiag{\Pfi^{-}}+\u\XX^{-}}$and
demanding stationarity of $S$ \eqref{eq:def_S_phi} with respect
to $\Pfi$  we obtain:

\begin{equation}
\begin{array}{cccrcccc}
0\overset{!}{=}\frac{\partial S}{\partial\Pfi}\:\Leftrightarrow\: & \tPfi & = & \frac{1}{2}\,\v\inv{\v\Pfi^{-}+\kap}\Bigl( & \YY & \:-\: & 0 & \Bigr)\inv{\v\Pfi^{-}+\kap}\\
 &  &  & +\:\frac{1}{2}\,\w\inv{\w\mdiag{\Pfi^{-}}+\u\XX^{-}}\Bigl( & \HH & \:-\: & (\w\mdiag{\Pfi^{-}}+\u\XX^{-}) & \Bigr)\inv{\w\mdiag{\Pfi^{-}}+\u\XX^{-}}\,,
\end{array}\label{eq:stat_Gam_C}
\end{equation}
..

This derivation yields the first main result of this work, cf. \eqref{eq:P_y__x}.

\subsubsection{Effect of partial temporal supervision}\label{subsec:Effect-of-partial}

\global\long\def\ve{\mathbf{e}}%
\global\long\def\vA{\mathbf{A}}%
\global\long\def\vU{\mathbf{U}}%
In the derivation above we assumed that the observation noise variance
$\kappa$ is the same for all timesteps. We now generalize to partial
temporal supervision, where only a subset of timesteps is observed.
The key insight is probabilistic: since $\kappa$ controls the observation
noise variance, setting $\kappa_{t}\rightarrow\infty$ at unobserved
timesteps $t$ makes the corresponding labels infinitely uncertain,
hence uninformative. Therefore, the label-dependent term in \prettyref{eq:stat_Gam_C}
should vanish at these points.

To verify this formally, let $\bar{\mathcal{T}}$ denote the set of
unobserved timesteps and define the projector $\vU\vU^{\T}=\sum_{t\in\mathcal{\bar{\mathcal{T}}}}\ve_{t}\ve_{t}^{\T}$
onto this subspace, where $\{\ve_{t}\}_{t=1\ldots T}$ is the standard
basis. The slack matrix becomes $\bm{\kappa}=\kappa(\mathbf{I}_{T}-\vU\vU^{\T})+\kappa_{\infty}\vU\vU^{\T}$.
For any matrix $\vA$ like the ones appearing in \prettyref{eq:stat_Gam_C},
the Woodbury identity gives
\begin{align*}
\left(\vA+\kappa_{\infty}\vU\vU^{\T}\right)^{-1}= & \vA^{-1}-\vA^{-1}\vU\inv{\kappa_{\infty}^{-1}\,+\,\vU^{\T}\vA^{-1}\vU}\vU^{\T}\vA^{-1}\\
\overset{\kap_{\infty}\rightarrow\infty}{\longrightarrow} & \vA^{-1}-\vA^{-1}\vU\inv{\vU^{\T}\vA^{-1}\vU}\vU^{\T}\vA^{-1}.
\end{align*}
This expression has vanishing support on $\text{span}\{\ve_{t}\}_{t\in\mathcal{\bar{\mathcal{T}}}}$,
as can be verified by multiplying with $\vU$ from the left and $\vU^{\T}$
from the right. Thus the immediate 'force' in the first line of \prettyref{eq:stat_Gam_C}
due to labels vanishes at unobserved timesteps, and only the NNGP
prior (second line) remains. Note that due to indirect feedback through
observed points, the posterior $\vev{hh^{\T}}$ will still deviate
from the NNGP prior; see \citet{Fischer24_10761}.

\subsubsection{Reduction to DNN case}\label{subsec:Reduction-to-DNN}

In their standard formulation, DNNs differ from the general architecture
developed in \prettyref{sec:Derivation-of-network-prior} in three
aspects:
\begin{enumerate}
\item There is only input in the first layer, $\bx^{t}=\delta^{t0}\bx^{0}$.
\item There is only output supervision on the last layer. This does not
amount to just setting $y^{t<T}=0,$ instead, we loosen the slack
in all but the last layer. This can also be interpreted as believing
that hypothetical evidence $y^{t<T}$ is non-salient (i.e., measurements
that have been corrupted by uninformative noise $\xi^{t<T}\sim\N(0,\,\kap_{\infty}\mathbf{I}^{t<T})$,
$\kap_{\infty}\gg\kap$). 
\item The weights under the prior are pairwise independent random variables
across layers, $P(\bW^{(t)},\bW^{(t+1)})=\N_{\bW}(\bW^{(t)})\N_{\bW}(\bW^{(t+1)})$.
After marginalization, this amounts to masking all off-diagonal elements
to be zero in the relevant expressions, denoted as $\mdiag{\Pfi}\rightarrow\text{diag}(\Pfi)$
(the vectorized version of replacing $\bar{\delta}^{tt'}\rightarrow\delta^{tt'}$),
i.e. only sampling the measure in the covariance that is the diagonal
of $\Pfi$.
\end{enumerate}
This leads to the following effect on the saddle-point equations,.
The kernel order parameter reads

\begin{equation}
\begin{array}{ccc}
\Pfi^{tt'} & = & \frac{\partial\cW}{\partial\tPfi^{tt'}}(\Pfi;\tPfi)=\vev{\phi(h^{t})\phi(h^{t'})}_{P(h|\Pfi,\tPfi,\YY,\XX)}\end{array}\,.\label{eq:C_app}
\end{equation}
The supremum condition for $\tPfi$ in \eqref{eq:def_gamma} becomes
\begin{equation}
\begin{array}{cccrcccc}
0\overset{!}{=}\frac{\partial S}{\partial\Pfi}\quad\Leftrightarrow\quad & \tPfi & = & \frac{1}{2}\v\inv{\v\Pfi^{-}+\kap}\Bigl( & \YY & \:-\: & 0 & \Bigr)\inv{\v\Pfi^{-}+\kap}\\
 &  &  & +\frac{1}{2}\w\inv{\w\text{diag}(\Pfi^{-})+\u\XX^{-}}\Bigl( & \HH & \:-\: & (\w\text{diag}(\Pfi^{-})+\u\XX^{-}) & \Bigr)\inv{\w\text{diag}(\Pfi^{-})+\u\XX^{-}}.
\end{array}\label{eq:tC_app}
\end{equation}

The first line can be interpreted as a tilt $\tPfi_{y}$ due to the
($\kappa$-regularized) constraint to match the labels $y$, and the
second line as a tilt $\tPfi_{h}$ due to the definition of the network's
forward pass. 

Due to the diagonal-only coupling, this set of equations can be reduced.
To this end, we make the diagonal Ansatz $\Pfi=\diag{\Pfi}$, $\tPfi=\diag{\tPfi}$,
$\HH=\diag{\HH}$. Then, the conjugate saddle point equation \prettyref{eq:C_app}
becomes

\begin{equation}
\begin{array}{cccrccccc}
 & \tPfi^{TT} & =\frac{1}{2} & \v\inv{\v\Pfi^{\Tm\Tm}+\kap}\Bigl( & \YY^{{\scriptscriptstyle TT}} & \:-\: & 0 & \Bigr)\inv{\v\Pfi^{\Tm\Tm}+\kap}\,,\\
 & \tPfi^{tt} & = & \frac{1}{2}\w\inv{\w\Pfi^{t-1,t-1}}\Bigl( & \vev{h^{t}\,h^{t\,\T}} & \:-\: & \w\Pfi^{t-1,t-1} & \Bigr)\inv{\w\Pfi^{t-1,t-1}}\,, & 2\leq t<T-1\\
 & \tPfi^{11} & = & \frac{1}{2}\w\inv{\u\XX^{0,0}}\Bigl( & \vev{h^{1}\,h^{1\,\T}} & \:-\: & \u\XX^{0,0} & \Bigr)\inv{\u\XX^{0,0}}\,,
\end{array}\label{eq:Ctilde_layer_fact}
\end{equation}
where we used $\kappa^{\leq T}=\kappa_{\infty}\rightarrow\infty$.
\prettyref{eq:Ctilde_layer_fact} reproduces the saddle-point equations
previously derived for DNNs \citet{Fischer24_10761,lauditi2025}.

\subsection{Special case of linear recurrent networks}\label{sec:linear}

We here consider the special case of a linear activation function
$\phi(h)=h$, in which the effective probability and saddle point
equations simplify considerably. 

\subsubsection{Kernel mean-field theory for the linear case}\label{subsec:Gamma_lin}

We derived the rate function \prettyref{eq:def_gamma} and the cumulant-generating
function \prettyref{eq:def_calW}, which in the linear case specialize
to

\[
P(\HH|\XX)\propto\exp\left\{ -N\Gamma(\HH|\XX)\right\} ,\quad\Gamma(\HH|\XX))=\sup_{\tPfi}\left(\tr\,\tPfi^{\T}\HH-\cW(\tPfi|\XX)\right)\,,
\]
\[
\cW(\tPfi|\XX)=\ln\int dh\,\exp\left\{ -\hlf h^{\T}\inv{\w\HH^{-}+\u\XX^{-}}h+h^{\T}\tPfi\,h-\hlf N\lndet{\w\HH^{-}+\u\XX}\right\} .
\]
Using the supremum condition for $\tPfi$ in \prettyref{eq:def_gamma}
\[
0\overset{!}{=}\frac{d}{d\tPfi}\,\left(\tPfi^{\T}\HH-\cW(\tPfi|\XX)\right)\,\Leftrightarrow\,\tPfi(\HH)=\frac{1}{2}\,\inv{\w\HH^{-}+\u\XX^{-}}-\frac{1}{2}\,\rinv{\HH},
\]
we get by plugging in

\begin{equation}
\begin{array}{ccccccc}
\Gamma(\HH|\XX) & = & \tr\,\tPfi(\HH)^{\T}\HH-\cW(\tPfi(\HH)|\XX)\\
 & = & \hlf\tr\,[\HH\left(\inv{\w\HH^{-}+\u\XX^{-}}-\rinv{\HH}\right)]\quad & -\ln\int dh\,\exp\biggl\{ & -\hlf h^{\T}\inv{\w\HH^{-}+\u\XX^{-}}h & +h^{\T}\tPfi h & -\hlf\lndet{\w\HH^{-}+\u\XX^{-}}\biggr\}\\
 & = & \hlf\tr\,[\HH\inv{\w\HH^{-}+\u\XX^{-}}-\I] & -\ln\int dh\,\exp\biggl\{ & -\hlf h^{\T}\inv{\w\HH^{-}+\u\XX^{-}}h & +\hlf h^{\T}\left(\inv{\w\HH^{-}+\u\XX^{-}}-\rinv{\HH}\right)h & -\hlf\lndet{\w\HH^{-}+\u\XX^{-}}\biggr\}\\
 & = & \hlf\tr\,[\HH\inv{\w\HH^{-}+\u\XX^{-}}-\I] & -\ln\int dh\,\exp\biggl\{ & -\hlf h^{\T}\rinv{\HH}h &  & -\hlf\lndet{\w\HH^{-}+\u\XX^{-}}\biggr\}\\
 & \overset{+\text{const.}}{=} & \hlf\tr\,[\HH\inv{\w\HH^{-}+\u\XX^{-}}]\quad &  &  &  & -\hlf\ln\,\frac{\det{\HH}}{\det{\w\HH^{-}+\u\XX^{-}}}
\end{array}\,,\label{eq:Gamma_lin}
\end{equation}
where we note that $\hlf\tr\,[\HH\inv{\w\HH^{-}+\u\XX^{-}}]-\hlf\ln\,\frac{\det{\HH}}{\det{\w\HH^{-}+\u\XX^{-}}}=D_{\text{KL}}\left(\N_{h}(0,\HH)\,||\,\N_{h}(0,\w\mdiag{\HH^{-}}+\u\XX^{-})\right)$
is the Kullback-Leibler divergence between two Gaussians in $h\in\bR^{T}$.
Note that this result can also be obtained directly from noticing
that the supremum condition on $\tPfi$ is made precisely such that
$\HH=\vev{hh^{\T}}$ (by definition of the Legendre transform). Hence
the integral over $h$ is a Gaussian with covariance $\HH$. 

\subsubsection{Closed-form solution}

Similarly to previous work, we can devise a closed-form solution for
the kernels by eliminating $\tPfi$. The basic reason this is possible
is that per \prettyref{eq:tC_app} 
\begin{align*}
\tPfi & =\frac{1}{2}\w\inv{\w\mdiag{\HH}+\u\XX}\left([hh^{\T}]_{\Pfi,\tPfi^{+}}^{+}-(\w\mdiag{\HH}+\u\XX)\right)\inv{\w\mdiag{\HH}+\u\XX}\quad+\quad\frac{1}{2}\,\v\inv{\v\HH+\kap}\YY^{+}\inv{\v\HH+\kap},
\end{align*}
and we in turn know that at the saddle it holds that $\Pfi^{+}=[hh^{\T}]_{\Pfi,\tPfi^{+}}^{+}$
giving
\[
\tPfi=\frac{1}{2}\w\inv{\w\mdiag{\HH}+\u\XX}\left(\HH^{+}-(\w\mdiag{\HH}+\u\XX)\right)\inv{\w\mdiag{\HH}+\u\XX}\quad+\quad\frac{1}{2}\v\inv{\v\HH+\kap}\YY^{+}\inv{\v\HH+\kap}.
\]
Finally, for the linear case we can read-off $\Pfi$ directly as the
coefficient of $h$ in the action, giving $\inv{\w\mdiag{\HH^{-}}+\u\XX^{-}}-2\tPfi=\rinv{\HH}$.
This gives a \textbf{closed-form relation} for $\HH$
\begin{align}
\inv{\w\mdiag{\HH^{-}}+\u\XX^{-}}-\rinv{\HH}=\w & \inv{\w\mdiag{\HH}+\u\XX}\left(\HH^{+}-(\w\mdiag{\HH}+\u\XX)\right)\inv{\w\mdiag{\HH}+\u\XX}\label{eq:MAP-closed-form-lin}\\
 & +\v\inv{\v\HH+\kap}\YY^{+}\inv{\v\HH+\kap}\nonumber 
\end{align}
where the last layer's term $2\tPfi_{h}^{{\scriptscriptstyle TT}}=\HH^{{\scriptscriptstyle TT}}-\w\Pfi^{\Tm\Tm}=0$
because the last layer is free to vary marginally.

This result of course corresponds to the stationary point of \prettyref{eq:def_S_phi}
in the special case of using \prettyref{eq:Gamma_lin}
\begin{align*}
S(\HH|\YY,\XX) & =-\hlf\,\tr\,[\YY\inv{\v\HH^{-}+\kap}]-\hlf\tr\,[\HH\inv{\w\HH^{-}+\u\XX^{-}}]+\hlf\ln\frac{|\HH|}{|\w\HH^{-}+\u\XX^{-}|}.
\end{align*}

\subsubsection{Perturbative solution}\label{subsec:Perturbative-solution}

We here derive the perturbative to solution to \prettyref{eq:MAP-closed-form-lin}
that we used in \prettyref{subsec:efficient-seq-learners} in the
main text. For the NNGP solution (i.e., absence of learning signal
$\YY$), we have $\HH_{0}=\w\mdiag{\HH_{0}^{-}}+\u\XX^{-}$. We then
make a perturbation expansion $\HH=\HH_{0}+\DD$, $\DD=\DD_{1}+\DD_{2}+\dots$,
where every order groups powers of $\YY$. We perform this expansion
up to second order, which is the minimal order at which the architecture-dependent
effects that we are interested in appear. Thus, we insert the perturbative
ansatz into \prettyref{eq:MAP-closed-form-lin} to get the expression
\[
\inv{\w(\HH_{0}+\DD)^{-}+\u\XX^{-}}\;-\;\inv{\HH_{0}+\DD}=\w\,\inv{\w\mdiag{\HH_{0}+\DD}+\u\XX}\,(\DD^{+}-\w\mdiag{\DD})\,\inv{\w\mdiag{\HH_{0}+\DD}+\u\XX}+\v\,\inv{\v(\HH_{0}+\DD)+\kap}\,\YY^{+}\,\inv{\v(\HH_{0}+\DD)+\kap}
\]

and group terms in zeroth, first, and second order in $\YY$ to compare
coefficients. 

\paragraph{Zeroth order}

The terms of zeroth order
\[
\inv{\w\HH_{0}^{-}+\u\XX^{-}}\;-\;\rinv{\HH_{0}}=0
\]
are fulfilled by having chosen the NNGP solution as the expansion
point, for which $\mdiag{\HH_{0}}=\HH_{0}$.

\paragraph{First order}

To ease notation, we introduce the NNGP ``propagators'' $\GG_{h}\coloneqq\inv{\w\HH_{0}^{-}+\u\XX^{-}}\equiv\inv{\HH_{0}}$,
$\GG_{h}^{+}\coloneqq\inv{\w\HH_{0}+\u\XX}$, $\GG_{y}\coloneqq\inv{\v\HH_{0}+\kap}$.
Expanding the LHS resolvents and the RHS linear terms yields the defining
equation for $\DD_{1}$:
\[
-(\GG_{h}\w\mdiag{\DD_{1}^{-}}\GG_{h}-\GG_{h}\DD_{1}\GG_{h})=\w\GG_{h}^{+}\left(\DD_{1}^{+}-\w\mdiag{\DD_{1}}\right)\GG_{h}^{+}\:+\:\v\GG_{y}\YY^{+}\GG_{y}\,,
\]
where we used $\mdiag{\HH_{0}}=\HH_{0}$ because the NNGP solution
is diagonal \citet{Segadlo22_accepted}. This moreover reveals that
the first order change is proportional to $\YY$.

\paragraph{Second order}

Grouping second-order terms $(\DD_{1})^{2}$, $\DD_{2}$, we get

\begin{align*}
-(\GG_{h}\w\mdiag{\DD_{2}^{-}}\GG_{h}-\GG_{h}\DD_{2}\GG_{h}) & = & \w\GG_{h}^{+}(\DD_{2}^{+}-\w\mdiag{\DD_{2}})\GG_{h}^{+}\\
+\GG_{h}\w\mdiag{\DD_{1}^{-}}\GG_{h}\w\mdiag{\DD_{1}^{-}}\GG_{h}-\GG_{h}\DD_{1}\GG_{h}\DD_{1}\GG_{h} &  & -\w\bigl(\GG_{h}^{+}\w\mdiag{\DD_{1}}\GG_{h}^{+}(\DD_{1}^{+}-\w\mdiag{\DD_{1}})\GG_{h}^{+}\:+\:(\sim)^{\T}\bigr)\\
 &  & -\v\bigl(\GG_{y}\v\DD_{1}\GG_{y}\YY^{+}\GG_{y}\:+\:(\sim)^{\T}\bigr)\,,
\end{align*}
where $(\sim)^{\T}$ denotes the transpose of the preceding term.
Notably, in the second line, masked $\mdiag{\DD_{1}}$ and unmasked
$\DD_{1}$ terms appear in a product, giving rise to the contractions
in the main text that allow for signal propagation.

\subsubsection{Solution degeneracy in linear RNNs}

Within the temporally-coherent phase of endpoint-supervised tasks,
there can technically also be other solutions, for example where the
hidden layer activity flips sign from one timestep to the next, $\bh^{t+1}=-\bh^{t}$.
Such solutions are unstable, since they would be suppressed by an
infinitesimal residual pathway (memory term) in \prettyref{eq:weight-model-RNN},
$\bh^{t+1}=\ldots+(1-\alpha)\bh^{t}$. To exclude this solution, we
start learning from a small $\alpha$, which we subsequently anneal
to zero.

\subsection{Analytics for temporally coherent phase in 4-layer linear RNNs in
terms of label strength}\label{subsec:Landau_L4}

\subparagraph{Setup}

We consider the case $T=4$ with $\kappa=0$. Let $\HH_{0:4}\in\mathbb{R}^{4\times4}$
be the kernel and $\XX_{0:4}^{-}\in\mathbb{R}^{4\times4}$ encode
input correlations. Here, we are adopted Python-like slicing conventions
that include the first and exclude the last index of a slice, i.e.
$1\!:\!4\coloneqq(1,2,3)$, and will put temporal indices into \textit{sub}scripts
to avoid confusion with exponents. We work in the endpoint-supervised
setting where in addition input is only supplied to the first layer,
so 
\begin{equation}
\XX^{-}=\mathrm{diag}(1,0,0,0)=\mathbf{e}_{0}\mathbf{e}_{0}^{\T}\,.\label{eq:input_kernel}
\end{equation}
Moreover, we consider a single training sample, i.e. $P=1$.

The log-probability (effective action) from \prettyref{eq:def_S_phi}
using \eqref{eq:Gamma_lin} for the end-point supervised task with
label $y_{4}$ can be written as 
\begin{equation}
\ell(\HH)\;\coloneqq\;\ln P(\HH|y,\bx)/N\;=-\frac{1}{2}\,\big[y_{4}^{2}\,\inv{\v\HH^{\Tm\Tm}+\kappa}\big]-\frac{1}{2}\,\mathrm{tr}\![\HH\inv{\Sigma_{h}(\HH)}]+\frac{1}{2}\,\ln\!\frac{\det{\HH}}{\det{\Sigma_{h}(\HH)}}\,,\label{eq:ell-general-1}
\end{equation}
where we introduced the shorthand $\Sigma_{h}(\HH)=\w\HH^{-}+\u\XX^{-}$
and used that, due to the supervision of the end point only, the probability
of the label $y_{4}$ is with regard to the marginal distribution
in the last layer alone (this is consistent with the consideration
in \prettyref{subsec:Effect-of-partial}, sending the regularization
noise $\kappa\to\infty$ for all unobserved timesteps). We will use
the dimensionless control parameter 
\begin{equation}
\lambda\;\coloneqq\;\frac{y_{4}^{2}}{\u\,\w^{2}\,\v}.\label{eq:lambda-def}
\end{equation}

\subparagraph{Parameterization of the relevant $3\times3$ block}

In the $T=4$ case, the nontrivial dependence of $\ell$ on off-diagonal
elements is confined to the lower-right $3\times3$ block of $\HH$,
corresponding to times $1,2,3$. We denote this block by $\HH_{1:4}\in\bR^{3\times3}$
and explicitly parametrize it as 
\[
\HH_{1:4}=\begin{bmatrix}a & b_{2} & b_{3}\\
b_{2} & c & d\\
b_{3} & d & e
\end{bmatrix},\quad\HH_{1:3}\coloneqq\begin{bmatrix}a & b_{2}\\
b_{2} & c
\end{bmatrix},\quad\HH_{2:4}=\begin{bmatrix}c & d\\
d & e
\end{bmatrix}.
\]

The observed output at the final time is $y_{4}$, so that $\v\,\HH_{3,3}=\v e$.

For the reduced $3\times3$ problem, the latent covariance over the
last three times decomposes into a driven $1\times1$ block and an
interior $2\times2$ block. The only architecture-dependent difference
(RNN vs.\ DNN) is whether the interior block inherits the off-diagonal
$b_{2}$:
\begin{equation}
\Sigma_{h}(\HH)=\begin{cases}
\u\,\mathbf{e}_{0}\mathbf{e}_{0}^{\T}\;\oplus\;\w\,\HH_{1:3}=\left[\begin{smallmatrix}\u & 0 & 0\\
0 & \w a & \w b_{2}\\
0 & \w b_{2} & \w c
\end{smallmatrix}\right] & \text{RNN (unmasked)}\\[2pt]
\u\,\mathbf{e}_{0}\mathbf{e}_{0}^{\T}\;\oplus\;\w\,\mathrm{diag}(\HH_{1:3})=\left[\begin{smallmatrix}\u & 0 & 0\\
0 & \w a & 0\\
0 & 0 & \w c
\end{smallmatrix}\right] & \text{DNN (masked)}
\end{cases}\,,\label{eq:sigma_h_arch}
\end{equation}
where we used the input kernel \eqref{eq:input_kernel} and employed
$\mathbf{e}_{0}=(1,0,0)^{\T}$ in the reduced $3$-dimensional indexing.

\subparagraph{Diagonal first-order conditions at $b_{2}=b_{3}=d=0$}

On the purely diagonal ansatz $(b_{2},b_{3},d)=(0,0,0)$, the RNN
and DNN coincide. The saddle point conditions $\frac{\partial}{\partial\{a,c\}}\ell\big(\HH(a,c,e)\big)\stackrel{!}{=}0$
then imply 
\begin{equation}
a^{2}=\frac{\u}{\w}\,c,\qquad e=\frac{c^{2}}{a}.\label{eq:diag-rel}
\end{equation}
We denote these stationarity conditions as ``first-order-conditions''
(FOC) in the following. The $e$--FOC $\frac{\partial}{\partial e}\ell\big(\HH(a,c,e)\big)\stackrel{!}{=}0$
reduces to a scalar relation between $e$, $c$ and the label: 
\begin{equation}
\frac{1}{e}-\frac{1}{\w\,c}+\frac{y_{4}^{2}}{\v\,e^{2}}=0.\label{eq:e-foc}
\end{equation}
These diagonal relations will later be used to locate the critical
point in terms of $\lambda$.

\subparagraph{Architecture-dependent gradient and linear response of interior off-diagonals}

The first-order condition with respect to the interior off-diagonal
$b_{2}$ differs between the RNN and DNN because $\Sigma_{h}(\HH)$
in~\eqref{eq:sigma_h_arch} depends on $b_{2}$ only for the RNN.
Writing the derivative by the matrix $\Sigma_{h}$ as 
\[
\frac{\partial\ell}{\partial\Sigma_{h}}=\frac{1}{2}\,\Sigma_{h}^{-1}(\HH-\Sigma_{h})\Sigma_{h}^{-1},
\]
the chain-rule gives
\[
\partial_{b_{2}}\ell=(\HH^{-1})_{12}\;+\;\begin{cases}
\w\big(\Sigma_{h}^{-1}(\HH-\Sigma_{h})\Sigma_{h}^{-1}\big)_{23}, & \text{RNN},\\[2pt]
0, & \text{DNN}.
\end{cases}
\]
Here we used two properties: first, that the first term of $\ell(\HH)$
does not depend on $b_{2}$, because it is $-\frac{1}{2}\,\big[y_{4}^{2}\,\inv{\v\HH^{\Tm\Tm}+\kappa}\big]$,
which only depends on $e$. Second, the trace can be explicitly calculated
as $\mathrm{tr}\![\HH\inv{\Sigma_{h}(\HH)}]=\text{sum}(\left[\begin{smallmatrix}a & b_{2} & b_{3}\\
b_{2} & c & d\\
b_{3} & d & e
\end{smallmatrix}\right]\odot\left[\begin{smallmatrix}1/\u & 0 & 0\\
0 & \w c/\det{\HH_{1:3}} & -\w b_{2}/\det{\HH_{1:3}}\\
0 & -\w b_{2}/\det{\HH_{1:3}} & \w a/\det{\HH_{1:3}}
\end{smallmatrix}\right]$), where we leveraged a block-inversion formula for $\rinv{(\Sigma_{h}(\HH))}$
and used $\mathrm{tr}\![AB]=\sum_{ij}A_{ij}B_{ij}\eqqcolon\text{sum}(A\odot B)$;
thus the $b_{2}$ dependence through $\HH$ in the first term vanishes.

Thus only in the RNN is there an additional feedback term driven by
the mismatch $\HH-\Sigma_{h}$.

To analyze the onset of an interior off-diagonal $d$ near the diagonal
point $(b_{2},b_{3},d)=(0,0,0)$, we treat the off-diagonal entries
$(d,b_{2},b_{3})$ as small and use Schur-complement identities for
$\HH^{-1}$. A direct calculation gives 
\[
(\HH^{-1})_{13}=\frac{-b_{2}d+b_{3}c}{\det{\HH}}.
\]
The FOC $\partial_{b_{3}}\ell=(\HH^{-1})_{13}=0$ therefore enforces
\[
b_{3}=\frac{d}{c}\,b_{2},
\]
so $b_{3}$ is slaved to $b_{2}$ and $d$ in a neighborhood of the
diagonal manifold.

We now parameterize the linear response of $b_{2}$ to $d$ by a coefficient
$\alpha$, 
\[
b_{2}=\alpha\,d+O(d^{3}),
\]
and determine $\alpha$ by enforcing the $b_{2}$--FOC. Substituting
$b_{3}=(d/c)\,b_{2}$ and $b_{2}=\alpha d$ into $\partial_{b_{2}}\ell=0$
and expanding to leading order in $d$ yields 
\[
\alpha=\begin{cases}
\dfrac{ac}{c^{2}+ae}, & \text{RNN},\\[6pt]
0, & \text{DNN}.
\end{cases}
\]
Consequently, near the diagonal manifold, 
\[
b_{2}=\alpha d+O(d^{3}),\qquad b_{3}=\frac{\alpha}{c}\,d^{2}+O(d^{4})\quad\text{(RNN)},\qquad b_{2}=b_{3}=0\quad\text{to these orders (DNN)}.
\]

\subparagraph{Quadratic curvature along the interior off-diagonal}

We now substitute 
\[
b_{2}=\alpha d,\qquad b_{3}=\frac{d}{c}\,b_{2}
\]
back into $\ell$ and expand in powers of $d$. The resulting Landau
expansion has the form 
\begin{equation}
\ell(d)=\ell_{0}+\frac{1}{2}\,C^{(2)}(\alpha)\,d^{2}-\frac{1}{4}\,\frac{d^{4}}{c^{2}e^{2}}+\cdots\,,\label{eq:landau}
\end{equation}
with 
\begin{equation}
C^{(2)}(\alpha)=-\frac{1}{ce}+\frac{\alpha}{\w ac}\quad\Rightarrow\quad\boxed{C_{\RNN}^{(2)}=-\frac{1}{ce}+\frac{1}{\w\,(c^{2}+ae)},\qquad C_{\text{DNN}}^{(2)}=-\frac{1}{ce}\ (<0).}\label{eq:C2-RNN-DNN}
\end{equation}
The term $-\frac{1}{ce}$ is the entropic penalty from $\ln\det{\HH}$
(Hadamard inequality: at fixed diagonal, the determinant $\det{\HH}$
is maximized when off-diagonals vanish). The quartic coefficient $-1/(4c^{2}e^{2})$
is architecture-independent and makes $-\ell$ locally stabilizing.

\subparagraph{Critical diagonals for the RNN}

A continuous transition occurs when the quadratic coefficient $C_{\RNN}^{(2)}$
changes sign. The critical point is defined by 
\[
C_{\RNN}^{(2)}=0\quad\Longleftrightarrow\quad\w(c^{2}+ae)=ce.
\]
Combining this with the diagonal relations~\eqref{eq:diag-rel},
\[
a^{2}=\frac{\u}{\w}\,c,\qquad e=\frac{c^{2}}{a},
\]
one obtains the criticality condition for three variables 
\begin{equation}
\boxed{a^{*}=2\,\u,\qquad c^{*}=4\,\u\w,\qquad e^{*}=8\,\u\w^{2}.}\label{eq:crit-diags}
\end{equation}
At $(a^{*},c^{*},e^{*})$ the RNN is marginal in the direction of
the interior off-diagonal $d$. In contrast, in the DNN one always
has $C_{2}^{\mathrm{DNN}}<0$, so the mode remains strictly massive
and no such transition occurs.

\subparagraph{Exact critical label strength in terms of $\lambda$}

Plugging~\eqref{eq:crit-diags} into the $e$--FOC~\eqref{eq:e-foc}
gives at criticality 
\[
\frac{1}{e^{*}}-\frac{1}{\w c^{*}}+\frac{y_{4}^{*\,2}}{\v\,(e^{*})^{2}}=0.
\]
Using $c^{*}=4\u\w$ and $e^{*}=8\u\w^{2}$, one has $1/(\w c^{*})=2/e^{*}$,
so the bracket simplifies to 
\[
-\frac{1}{e^{*}}+\frac{y_{4}^{*\,2}}{\v\,(e^{*})^{2}}=0\quad\Longleftrightarrow\quad y_{4}^{*\,2}=\v e^{*}=8\,\u\,\v\,\w^{2}.
\]
Equivalently, in terms of the dimensionless control parameter~\eqref{eq:lambda-def},
\begin{equation}
\boxed{\lambda^{*}=\frac{y_{4}^{*\,2}}{\u\,\w^{2}\,\v}=8.}\label{eq:crit-lambda}
\end{equation}
Since the label enters only through $y_{4}^{2}$, the sign of $y_{4}$
is irrelevant for the transition. For $\lambda<8$, one has $C_{\RNN}^{(2)}<0$
and the unique maximizer is $d^{*}=0$; for $\lambda>8$, one has
$C_{\RNN}^{(2)}>0$ and two symmetric non-zero maximizers $\pm d^{*}\neq0$
appear.

\subparagraph{Scaling near criticality without auxiliary variables}

Eliminating the diagonal displacement in favor of $\lambda-8$ gives
a linear expansion 
\[
C^{(2)}(\lambda)=K_{\lambda}\,(\lambda-8)+o(\lambda-8),\qquad K_{\lambda}=\frac{1}{1280\,\u^{2}\,\w^{3}}.
\]
Maximizing $\ell(d)$ in~\eqref{eq:landau} away from $d=0$ yields
\[
d^{*\,2}=C^{(2)}\,c^{2}e^{2}.
\]
Evaluating $c^{2}e^{2}$ at the critical diagonals~\eqref{eq:crit-diags}
gives $c^{*\,2}e^{*\,2}=1024\,\u^{4}\w^{6}$, hence 
\begin{equation}
\boxed{d^{*\,2}=\frac{4}{5}\,\u^{2}\,\w^{3}\,(\lambda-8)+o(\lambda-8)\qquad(\lambda\downarrow8^{+}).}\label{eq:dstar2-lambda}
\end{equation}
Equivalently, $d^{*}=\pm\sqrt{\frac{4}{5}\u^{2}\w^{3}(\lambda-8)}+o\!\big(\sqrt{\lambda-8}\big)$,
i.e.\ the usual mean-field exponent $\beta=\tfrac{1}{2}$ for $d$.

For the concrete choice $\u=\v=\w=1$, one has $\lambda=y_{4}^{2}$,
$\lambda^{*}=8$, and 
\[
d^{*\,2}=\frac{4}{5}\,(\lambda-8)+o(\lambda-8).
\]

\subsection{Numerical solution of kernels for saddle point equations}

We here describe how to obtain the maximum a posteriori kernels that
describe converged training for the linear kernels $\HH$ \prettyref{eq:P_H_lin}
and nonlinear kernels $\Pfi$, $\tPfi$ \prettyref{eq:P_y__x}. For
both solvers, we initialize the kernels at the NNGP solution and add
a slight off-diagonal for $\HH$ and $\Pfi$ to break symmetry.

\paragraph{Linear case}

We use SciPy's Newton-CG optimizer, a second-order method that leverages
Hessian-vector products for efficient curvature approximation without
forming the full Hessian. Gradients and Hessian-vector products are
computed via automatic differentiation in JAX.

\paragraph{Nonlinear case}

The GD solver iteratively refines the kernels $(\Pfi,\tilde{C})$
by computing gradients of the free energy $S\coloneqq\ln P(\Pfi,\tilde{C})$
through two passes: a ``\textit{forward pass}'' that evaluates the
feature moments at the \textit{equilibrium} kernels $(\HH_{\text{eql}}^{+}=\vev{h^{+}h^{+\T}}_{(\Pfi,\tilde{C})},\,\Pfi_{\mathrm{eql}}^{+}=\vev{\phi(h^{+})\phi(h^{+})^{\T}}_{(\Pfi,\tilde{C})})$
under a tilted Gaussian measure parameterized by the current solver
state $(\Pfi,\tilde{C})$, and a \textit{``backward pass}'' that
computes the equilibrium dual kernels $\tilde{C}_{\mathrm{eql}}$
via matrix inversions. The gradients take the symmetric form 
\[
\nabla_{\Pfi}S=\tPfi-\tilde{C}_{\mathrm{eql}},
\]
\[
\nabla_{\tilde{C}}S=\Pfi-\Pfi_{\mathrm{eql}}.
\]
We perform gradient \textit{descent} in $\Pfi$, projecting the gradient
to preserve positive definiteness, and gradient \textit{ascent} in
$\tilde{C}$. We perform these updates using an extragradient scheme
that recomputes gradients at half steps. 

\subsection{Overview table of mean-field derivation}\label{subsec:Summary-mft-derivation}

In this section, we give a compact overview of the derivation in \prettyref{sec:Derivation-of-network-prior}
to give a clearly traceable route through the derivation, with definitions
given there. In particular, this will reveal why the chosen intensive
scaling of the prior variances yields a well-defined set of saddle
point equations, explaining why the chosen order parameters concentrate.
We will adopt a ``greedy'' version of Einstein summation convention,
where summations should be placed as tightly as possible around repeated
indices. We will here not explicitly write out contractions over indices
$p=1\ldots P$ for clarity. We will also assume $P,T\ll N$ and drop
any such subleading terms. 

Below we make a step-by-step derivation of the network's prior predictive
distribution \eqref{eq:network_prior} 
\begin{align*}
P(y|\bx) & =\int d\bU d\bW\,d\bV\;P(\bU)\,P(\bW)\,P(\bV)\\
 & \times\int d\bh\,d\xi\;P(\xi^{t})\;\prod_{t}^{T}P(y^{t}\mid\bh^{t-1},\bV,\xi^{t})\;\prod_{t}^{T}P(\bh^{t}\mid\bh^{t-1},\bx^{t-1},\bU,\bW)\,,
\end{align*}
where $\bh\in\bR^{\Tm P\times N}$, $\xi\in\bR^{TP}$.

{ %
\makeatletter
\let\orig@hline\hline
\renewcommand{\hline}{%
  \noalign{\vskip 0.6ex}%
  \orig@hline
  \noalign{\vskip 0.6ex}%
}
\makeatother

\begin{adjustbox}{width=\textwidth,left}
\setlength{\arraycolsep}{0.5pt}
\begin{minipage}{\linewidth}

\begin{align*}
\begin{array}{clccccccclcccc}
 & P(y|\bx)\\
= & \int_{\begin{array}{c}
\bh\xi\end{array}} & \int_{\bU\bW\bV} &  &  &  &  &  & \delta\Bigl( & y^{t} & -\:(\bV\bphi^{t-1} & +\xi^{t-1}) &  & \Bigr)\\
 &  &  &  &  &  &  &  & \delta\Bigl( & \bh^{t} & -\:(\bW^{(t-1)}\bphi^{t-1} & +\bU\bx^{t-1}) &  & \Bigr)\\
 &  &  &  &  &  &  &  & \N_{\bU}(\bU) & \N_{\bW}(\{\bW^{(t)}\}_{t}) & \N_{\bV}(\bV) & \N_{\Kap}(\xi)\\
\hline \overset{1.}{=} & \int_{\bh\xi\tbh\tf} & \int_{\bU\bW\bV} &  &  &  &  &  & \exp\Bigl\{ & \i\tf^{t}y^{t} & -\i\tf^{t}\bV\bphi^{t-1} & -\i\tf^{t}\xi^{t} &  & \Bigr\}\\
 &  &  &  &  &  &  &  & \exp\Bigl\{ & \i\tbh^{t}\bh^{t} & -\i\tbh^{t}\bW^{(t-1)}\bphi^{t-1} & -\i\tbh^{t}\bU\bx^{t-1} &  & \Bigr\}\\
 &  &  &  &  &  &  &  & \N_{\bU}(\bU) & \N_{\bW}(\{\bW^{(t)}\}_{t}) & \N_{\bV}(\bV) & \N_{\Kap}(\xi)\\
\hline \overset{2.}{=} &  &  &  &  &  & \int_{\bh\tbh\tf} &  & \exp\Bigl\{ & \i\tf^{t}y^{t} & -\hlf\tf^{t}(V\phi_{i}^{t-1}\phi_{i}^{t'-1})\tf^{t'} & -\hlf\tf^{t}\Kap\sdelta^{tt'}\tf^{t'}\\
 &  &  &  &  &  &  &  &  & \i\th_{i}^{t}h_{i}^{t} & -\th_{i}^{t}(W\bsdelta^{ts}\phi_{j}^{t-1}\phi_{j}^{t'-1})\th_{i}^{t'} & -\hlf\th_{i}^{t}(Ux_{j}^{t-1}x_{j}^{t'-1})\th_{i}^{t'} &  & \Bigr\}\\
\hline \overset{3.}{=} & \int_{\Pfi\tPfi} & \exp\Bigl\{ & -\tPfi^{tt'}\Pfi^{tt'} & \Bigr\} &  & \int_{\bh\tbh\tf} &  & \exp\Bigl\{ & \i\tf^{t}y^{t} & -\hlf\tf^{t}(VN\Pfi^{t-1,t'-1})\tf^{t'} & -\hlf\tf^{t}\Kap\sdelta^{tt'}\tf^{t'}\\
 &  &  &  &  &  &  &  &  & \i\th_{i}^{t}h_{i}^{t} & -\frac{1}{N}\phi_{i}^{t}\tPfi^{tt'}\phi_{i}^{t'} & -\th_{i}^{t}(NW\enspace\bsdelta^{\,tt'}\Pfi^{t-1,t'-1})\th_{i}^{t'} & -\hlf N\th_{i}^{t}(DU\XX^{t-1,t'-1})\th_{i}^{t'} & \Bigr\}\\
\hline \overset{4.}{=} & \int_{\Pfi\tPfi} & \exp\Bigl\{ & -\tPfi^{tt'}\Pfi^{tt'} & \Bigr\} &  & \int_{\bh\tbh\tf} &  & \exp\Bigl\{ & \i N\tf^{t}y^{t} & -\hlf N^{2}\tf^{t}(VN\Pfi^{t-1,t'-1})\tf^{t'} & -\hlf N^{2}\tf^{t}\Kap\sdelta^{tt'}\tf^{t'}\\
 &  &  &  &  &  &  &  &  & \i\th_{i}^{t}h_{i}^{t} & -\frac{1}{N}\phi_{i}^{t}\tPfi^{tt'}\phi_{i}^{t'} & -\th_{i}^{t}(NW\enspace\bsdelta^{\,tt'}\Pfi^{t-1,t'-1})\th_{i}^{t'} & -\hlf N\th_{i}^{t}(DU\XX^{t-1,t'-1})\th_{i}^{t'} & \Bigr\}\\
\hline \overset{5.}{=} & \int_{\Pfi\tPfi} & \exp\Bigl\{ & -\tPfi^{tt'}\Pfi^{tt'} & \Bigr\} &  & \int_{\tf} &  & \exp\Bigl\{ & \i N\tf^{t}y^{t} & -\hlf N^{2}\tf^{t}(VN\Pfi^{t-1,t'-1})\tf^{t'} & -\hlf N^{2}\tf^{t}\Kap\sdelta^{tt'}\tf^{t'} &  & \Bigr\}\\
 &  &  &  &  & \times & \int_{\bh\tbh} & \prod_{i} & \exp\Bigl\{ & \i\th_{i}^{t}h_{i}^{t} & \frac{1}{N}\phi_{i}^{t}\tPfi^{tt'}\phi_{i}^{t'} & -\th_{t,i}(NW\enspace\bsdelta^{\,tt'}\Pfi^{t-1,t'-1})\th_{i}^{t'} & -\hlf N\th_{i}^{t}(DU\XX^{t-1,t'-1})\th_{i}^{t'} & \Bigr\}\\
\hline \overset{6.}{=} & \int_{\Pfi\tPfi} & \exp\Bigl\{ & -\tPfi^{tt'}\Pfi^{tt'} & \Bigr\} &  & \int_{\tf} &  & \exp\Bigl\{ & \i N\tf^{t}y^{t} & -\hlf N^{2}\tf^{t}(VN\Pfi^{t-1,t'-1})\tf^{t'} & -\hlf N^{2}\tf^{t}\Kap\sdelta^{tt'}\tf^{t'}\\
 &  &  &  &  & \times & \int_{\bh\tbh} & \prod_{i} & \exp\Bigl\{ & \i\th^{t}h^{t} & \phi^{t}\tPfi^{tt'}\phi^{t'} & -\th^{t}(NW\enspace\bsdelta^{\,tt'}\Pfi^{t-1,t'-1})\th^{t'} & -\hlf\th^{t}(DU\XX^{t-1,t'-1})\th^{t'} & \Bigr\}\\
\hline \overset{7.}{=} & \int_{\Pfi\tPfi} & \exp\Bigl\{ & -N\tPfi^{tt'}\Pfi^{tt'} & \Bigr\} &  & \int_{\tf} &  & \exp\Bigl\{ & \i N\tf^{t}y^{t} & -\hlf N\tf^{t}(\v\Pfi^{t-1,t'-1})\tf^{t'} & -\hlf N\tf^{t}\kappa\sdelta^{tt'}\tf^{t'}\\
 &  &  &  &  & + & N & \ln\Bigl[\int_{h\th} & \exp\bigl\{ & \i\th^{t}h^{t} & \phi^{t}\tPfi^{tt'}\phi^{t'} & -\th^{t}(\w\enspace\bsdelta^{\,tt'}\Pfi^{t-1,t'-1})\th^{t'} & -\hlf\th_{i}^{t}(\u\XX^{t-1,t'-1})\th_{i}^{t'} & \bigr\}\Bigr]\Bigr\}\\
\hline \overset{8.}{=} & \int_{\Pfi\tPfi} & \exp\Bigl\{ & -N\tPfi^{tt'}\Pfi^{tt'} & \Bigr\} &  &  &  & \exp\Bigl\{ &  & -\hlf y^{t}N\bigl(\inv{\v\Pfi^{-}+\kap}\bigr)^{tt'}y^{t'} &  & -\hlf\lndet{\fN(\v\Pfi^{-}+\kap)}\\
 &  &  &  &  & + & N & \ln\Bigl[\int_{h\th} & \exp\bigl\{ &  & \phi^{t}\tPfi^{tt'}\phi^{t'} & -\hlf h^{t}\bigl(\inv{\w\mdiag{\Pfi^{-}}+\u\XX^{-}}\bigr)^{tt'}h^{t'} & -\hlf\lndet{\w\mdiag{\Pfi^{-}}+\u\XX^{-}} & \Bigr\}\Bigr]\\
\hline \overset{9.}{=} & \int_{\Pfi\tPfi} & \exp\Bigl\{ & N\Bigl[-\tPfi^{tt'}\Pfi^{tt'} &  &  &  &  &  &  & -\hlf y^{t}\bigl(\inv{\v\Pfi^{-}+\kap}\bigr)^{tt'}y^{t'} &  & -\hlf\fN\lndet{\fN(\v\Pfi^{-}+\kap)}\\
 &  &  &  &  & - &  & \ln & Z(\tPfi;\Pfi) &  &  &  &  & \Bigr]\Bigr\}
\end{array}
\end{align*}

\end{minipage}
\end{adjustbox}
}

{} %

Herein, we made the following manipulations:
\begin{enumerate}
\item Rewrite each Dirac delta enforcing the deterministic network equations
in its Fourier representation, $\delta(\circ)=\int d\tilde{z}\,\exp\{\i\tilde{z}(\circ)\}$,
introducing conjugate fields $\tilde{f}^{t},\tilde{h}^{t}$ for the
outputs $y^{t}$ and hidden pre-activations $h^{t}$. This converts
the constraints into linear couplings in the exponent, such that all
random variables only appear in the equation through quadratic forms.
\item Integrate out the Gaussian weight priors $\{\bW^{(t)}\}_{t}$, $\bU$,
$\bV$ and the output noise $\xi$. We introduce the variable $\bsdelta^{tt'}$
\[
\bsdelta^{tt'}\coloneqq\begin{cases}
\delta^{tt'}, & \text{DNN}\\
1, & \text{RNN}
\end{cases}
\]
which allows us to write down the covariance of the priors in the
short form $\ev{\bW_{ij}^{(t)}\bW_{ij}^{(t')}}=W\,\bsdelta^{tt'}$
(and analogously for $U,V,\kappa$). We denote the corresponding masking
operator on time indices by 
\[
(\mdiag A)^{tt'}\coloneqq\bsdelta^{tt'}\,A^{tt'},
\]
which acts as a diagonal projector in $t,t'$ for DNNs and as the
identity for RNNs.
\item Introduce the empirical kernel $\Pfi^{tt'}\coloneqq\frac{1}{N}\sum_{i=1}^{N}\phi_{i}^{t}\phi_{i}^{t'}$
via a Dirac delta, $\delta\left(\Pfi-\frac{1}{N}\sum_{i}\phi_{i}\phi_{i}\right)=\int d\tPfi\,\exp\left\{ \tPfi^{tt'}\Bigl(\Pfi^{tt'}-\frac{1}{N}\sum_{i}\phi_{i}^{t}\phi_{i}^{t'}\Bigr)\right\} ,$in
order to decouple different neurons. The $\int d\tPfi$ integral is
taken along a contour parallel to the imaginary axis passing through
the saddle point; we omit an explicit imaginary unit, anticipating
that the saddle lies on the real axis. As long as there are no singularities
between the original and shifted contours, such deformations are justified
(Bromwich contour).
\item Rescale the output conjugate field according to $\tilde{f}\mapsto N\tilde{f}$
so that its quadratic form scales as $\mathcal{O}(N)$, on the same
footing as the contribution from the hidden conjugate fields. This
is an algebraic change of variables that prepares the expression for
a uniform large-$N$ analysis.
\item Exploit the fact that now, conditional on the collective fields $(\Pfi,\tPfi)$,
different neurons are independent and identically distributed. This
allows us to factor the $N$-dimensional integral over $\{h_{i},\tilde{h}_{i}\}_{i=1}^{N}$
into a product of identical single-site integrals.
\item Rescale the conjugate kernel to make the overall $N$-dependence explicit,
$\tPfi\mapsto N\,\tPfi$. After this redefinition the kernel term
in the exponent takes the canonical form $-\frac{N}{2}\tPfi^{tt'}\Pfi^{tt'}$,
so that the full effective action is proportional to $N$.
\item Insert the mean-field (‘intensive’) scalings for the prior variances,
e.g.\ $V=\nicefrac{\v}{N^{2}}$, $W=\nicefrac{\w}{N}$, $U=\nicefrac{\w}{N}$
(and similarly for the noise level), so that the pre-activation covariances
and outputs remain of order one as $N\to\infty$. This yields quadratic
forms involving $v,w,u$ and the kernels $\Pfi$ rather than the original
extensive parameters.
\item Perform the remaining Gaussian integrals over $\tilde{f}$ and over
the hidden fields (after integrating out their conjugate fields inside
the single-site factor). This produces the quadratic form in the outputs
$y$, with covariance $\v\Pfi^{-}+\kap$, as well as the corresponding
log-determinant normalization, and an analogous Gaussian contribution
for the hidden fields involving the matrix $\w\mdiag{\Pfi^{-}}+\u\XX^{-}$.
\item Recognize that the contribution of a single neuron is captured by
a partition function $Z(\tPfi;\Pfi)$ which only depends on the collective
fields through $(\Pfi,\tPfi)$. Because all neurons are identically
distributed, the full factor from the hidden layer is $Z(\tPfi;\Pfi)^{N}$,
which yields a term $-N\ln Z(\tPfi;\Pfi)$ in the effective action.
\item Apply a saddle-point approximation (Laplace’s method) to the remaining
functional integral over $(\Pfi,\tPfi)$ in the limit of large width
$N$. The exponent is of the form $N\,S(\Pfi,\tPfi)$, so the integral
is dominated by the stationary points of $S$; this is equivalent
to invoking a large-deviation principle for the empirical kernel.
\end{enumerate}
A key object that arises from this calculation is the single-site
cumulant generating function

\begin{equation}
W(\tPfi;\Pfi)=\ln Z(\tPfi;\Pfi)=\ln\int_{h\th}\exp\Bigl\{\,-\phi^{t}\tPfi^{tt'}\phi^{t'}-\hlf h^{t}\bigl(\inv{\w\mdiag{\Pfi^{-}}+\u\XX^{-}}\bigr)^{tt'}h^{t'}-\hlf\lndet{\w\mdiag{\Pfi^{-}}+\u\XX^{-}}\Bigr\}.\label{eq:W}
\end{equation}

whose derivatives with respect to $\tPfi$ yield the cumulants $\phi\phi^{\T}$.

\newpage{}

\subsection{Additional figures}\label{sec:Additional-figures}

\begin{figure}[H]
\centering
\centering{}\includegraphics[width=1\textwidth]{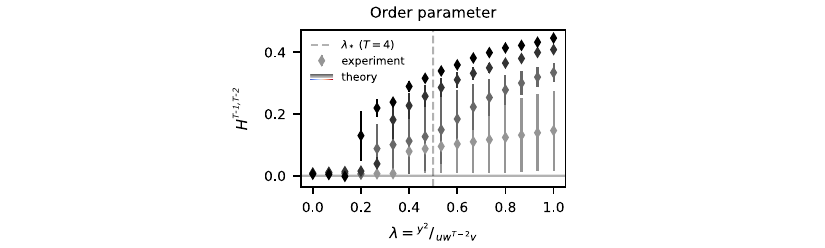}\caption{\textbf{Like \prettyref{fig:RNNs_vs_DNNs_endpoint_sv}, but with
nonlinear activation $\phi(\circ)=\protect\erf(\tfrac{\sqrt{\pi}}{2}\circ)$.}}\label{fig:RNNs_vs_DNNs_endpoint_sv_erf}
\end{figure}

\clearpage{} 
\end{document}